\documentclass[format=acmsmall, review=false, screen=true]{acmart}

\usepackage[lined,boxed,commentsnumbered]{algorithm2e}
\usepackage{amsmath}
\usepackage{booktabs} 
\usepackage{graphicx}
\usepackage{multirow}
\usepackage{rotating}
\usepackage{soul}
\usepackage{url}
\usepackage[utf8]{inputenc}
\usepackage{subcaption}
\usepackage[export]{adjustbox}
\usepackage{stackengine}
\usepackage{amsfonts}
\usepackage{amsthm}
\usepackage[normalem]{ulem}
\usepackage{xcolor}
\usepackage{colortbl}
\usepackage{subcaption}
\newcommand{\killpunct}[1]{}

\theoremstyle{definition}

\newif\ifdraft

\newcommand{\blue}[1]{\ifdraft{\leavevmode\color{blue}{#1}}\else{\leavevmode\color{black}{#1}}\fi}

\newcommand{\grayed}[1]{{\cellcolor[gray]{.7}}}

\newcommand{\fkfcv}{\textsc{Fun(kfcv)}}
\newcommand{\ftat}{\textsc{Fun(tat)}}
\newcommand{\fun}{\textsc{Fun}}
\newcommand{\gfun}{\textsc{gFun}}

\newcommand{\tfidf}{\mathrm{TFIDF}}
\newcommand{\tf}{\mathrm{TF}}

\newcommand{\sddag}{$^{\ddag}$}
\newcommand{\sdpsdag}{$^{\dag}$}
\newcommand{\psdag}{\phantom{$^{\dag}$}}

\newcommand{\TP}{\mathrm{TP}}
\newcommand{\TN}{\mathrm{TN}}
\newcommand{\FP}{\mathrm{FP}}
\newcommand{\FN}{\mathrm{FN}}

\newcommand{\vgx}{-X}
\newcommand{\vgm}{-M}
\newcommand{\vgw}{-W}
\newcommand{\vgxm}{-XM}
\newcommand{\vggM}{-R$_{\mathrm{M}}$}
\newcommand{\vggMW}{-R$_{\mathrm{MW}}$}
\newcommand{\vgb}{-B}
\newcommand{\vgmw}{-MW}

\newcommand{\vgxmb}{-XMB}
\newcommand{\vgxwb}{-XWB}
\newcommand{\vgxmw}{-XMW}
\newcommand{\vgwmb}{-WMB}
\newcommand{\vgbgMW}{-BR$_{\mathrm{MW}}$}
\newcommand{\vgxgMW}{-XR$_{\mathrm{MW}}$}
\newcommand{\vgxbgM}{-XBR$_{\mathrm{M}}$}
\newcommand{\vgxwmb}{-XWMB}
\newcommand{\vgxbgMW}{-XBR$_{\mathrm{MW}}$}
\newcommand{\vgbgM}{-BR$_{\mathrm{M}}$}

\copyrightyear{2021}
\acmYear{2021}
\setcopyright{acmlicensed}
\acmPrice{XX.XX}
\acmDOI{XX.XXXX/XXXXX.XXXX}
\acmISBN{XXXX}

\acmJournal{TOIS}

\begin{document}

\title[Generalized Funnelling]{Generalized Funnelling: \\ Ensemble
Learning and Heterogeneous Document Embeddings for Cross-Lingual Text
Classification} \thanks{The order in which the authors are listed is
purely alphabetical; each author has given an equally important
contribution to this work.}


\author{Alejandro Moreo} \authornotemark[1]
\email{alejandro.moreo@isti.cnr.it} \orcid{0000-0002-0377-1025}
\affiliation{%
\institution{Istituto di Scienza e Tecnologie dell'Informazione,
Consiglio Nazionale delle Ricerche, 56124 Pisa, Italy}
}

\author{Andrea Pedrotti} \orcid{0000-0002-2322-7043}
\email{andrea.pedrotti@phd.unipi.it}
\affiliation{%
\institution{Dipartimento di Informatica, Università di Pisa, 56127
Pisa, Italy}
}

\author{Fabrizio Sebastiani} \authornotemark[1]
\email{fabrizio.sebastiani@isti.cnr.it} \orcid{0000-0003-4221-6427}
\affiliation{%
\institution{Istituto di Scienza e Tecnologie dell'Informazione,
Consiglio Nazionale delle Ricerche, 56124 Pisa, Italy}
}

%
\renewcommand{\shortauthors}{Moreo, Pedrotti, Sebastiani}


\begin{abstract}
  \noindent \emph{Funnelling} (\fun) is a recently proposed method for
  cross-lingual text classification (CLTC) based on a two-tier
  learning ensemble for heterogeneous transfer learning (HTL). In this
  ensemble method, 1st-tier classifiers, each working on a different
  and language-dependent feature space, return a vector of calibrated
  posterior probabilities (with one dimension for each class) for each
  document, and the final classification decision is taken by a
  meta-classifier that uses this vector as its input. The
  meta-classifier can thus exploit class-class correlations, and this
  (among other things) gives \fun\ an edge over CLTC systems in which
  these correlations cannot be brought to bear. In this paper we
  describe \emph{Generalized Funnelling} (\gfun), a generalisation of
  \fun\ consisting of an HTL architecture in which 1st-tier components
  can be arbitrary \emph{view-generating functions}, i.e.,
  language-dependent functions that each produce a
  language-independent representation (``view'') of the
  \blue{(monolingual)} \label{line:monolingual2} document. We describe
  an instance of \gfun\ in which the meta-classifier receives as input
  a vector of calibrated posterior probabilities (as in \fun)
  aggregated to other embedded representations that embody other types
  of correlations, such as word-class correlations (as encoded by
  \emph{Word-Class Embeddings}), word-word correlations (as encoded by
  \emph{Multilingual Unsupervised or Supervised Embeddings}), and
  word-context correlations (as encoded by \emph{multilingual
  BERT}). We show that this instance of \textsc{gFun} substantially
  improves over \fun\ and over state-of-the-art baselines, by
  reporting experimental results obtained on two large, standard
  datasets for multilingual multilabel text classification. Our code
  that implements \gfun\ is publicly available.
\end{abstract}

%
%

%

\maketitle



\section{Introduction}
\label{sec:introduction}

\noindent \emph{Transfer Learning} (TL)~\cite{Vilalta:2011fk} is a
class of machine learning tasks in which, given a training set of
labelled data items sampled from one or more ``source'' domains, we
must issue predictions for unlabelled data items belonging to one or
more ``target'' domains, related to the source domains but different
from them. In other words, the goal of TL is to ``transfer'' (i.e.,
reuse) the knowledge that has been obtained from the training data in
the source domains, to the target domains of interest, for which few
labelled data (or no labelled data at all) exist. The rationale of TL
is thus to increase the performance of a system on a downstream task
(when few labelled data for this task exist), or to make it possible
to carry out this task at all (when no training data at all for this
task exist), while avoiding the cost of annotating new data items
specific to this task.

TL techniques can be grouped into two main categories, according to
the characteristics of the feature spaces in which the instances are
represented.
\emph{Homogeneous} TL (which is often referred to as \textit{domain
adaptation}~\cite{Zhang:2019fg}) encompasses problems in which the
source instances and the target instances are represented in a shared
feature space. Conversely, \emph{heterogeneous} TL~\cite{Day:2017qe}
denotes the case in which the source data items and the target data
items lie in different, generally non-overlapping feature spaces.
This article focuses on the heterogeneous case only; from now on, by
HTL we will thus denote \emph{heterogeneous} transfer learning.

A prominent instance of HTL in the natural language processing and
text mining areas is \textit{Cross-Lingual Transfer Learning} (CLTL),
in which data items have a textual nature and the different domains
are actually different languages in which the data items are
expressed. In turn, an important instance of CLTL is the task of
\emph{cross-lingual text classification} (CLTC), which consists of
classifying documents, each written in one of a finite set
$\mathcal{L}=\{\lambda_{1},...,$ $\lambda_{|\mathcal{L}|}\}$ of
languages, according to a shared \textit{codeframe} (a.k.a.\
\textit{classification scheme})
$\mathcal{Y}=\{y_{1}, ..., y_{|\mathcal{Y}|}\}$. The brand of CLTC we
will consider in this paper is (cross-lingual) \textit{multilabel}
classification, namely, the case in which any document can belong to
zero, one, or several classes at the same time. 

The CLTC literature has focused on two main variants of this task. The
first \blue{variant} (that is sometimes called the \emph{many-shot}
variant) deals with the situation in which the target languages are
such that language-specific training data are available for them as
well; in this case, the goal of CLTC is to improve the performance of
target language classification with respect to what could be obtained
by leveraging the language-specific training data alone. If these
latter data are few, the task if often referred to as \emph{few-shot}
learning. \blue{(We will deal with the
many-shot/few-shot scenario in the experiments of
Section~\ref{sec:baselines}.)} The second \blue{variant} is usually
called the \textit{zero-shot} variant, and deals with the situation in
which there are no training data at all for the target languages; in
this case, the goal of CLTC is to allow the generation of a classifier
for the target languages, which could not be obtained
otherwise. \blue{(We will deal with the
zero-shot scenario in the experiments of Section~\ref{sec:ZSCLC}.)}

\blue{Many-shot CLTC is important, since
in many multinational organisations (e.g., Vodafone, FAO, the European
Union) many labelled data may be available in several languages, and
there may be a legitimate desire to improve on the classification
accuracy that monolingual classifiers are capable of delivering. The
importance of few-shot and zero-shot CLTC instead} lies in the fact
that, while modern learning-based techniques for NLP and text mining
have shown impressive performance when trained on huge amounts of
data, there are many languages for which data are scarce.
According to~\cite{Joshi:2020my}, the amount of (labelled and
unlabelled) resources for the more than 7,000 languages spoken around
the world follows (somehow unsurprisingly) a power-law distribution,
i.e., while a small set of languages account for most of the available
data, a very long tail of languages suffer from data scarcity, despite
the fact that languages belonging to this long tail may have large
speaker bases.
Few-shot / zero-shot CLTL thus represents an appealing solution to
dealing with this situation, since it attempts to bridge the gap
between the high-resource languages and the low-resource
ones.

However, the application of CLTC is not necessarily limited to
scenarios in which the set of the source languages and the set of the
target languages are disjoint, nor it is necessarily limited to cases
in which there are few or no training data for the target domains.
CLTC can also be deployed in scenarios where a language can play both
the part of a source language (i.e., contribute to performing the task
in other languages) and of a target language (i.e., benefit from
training data expressed in other languages),
and where sizeable quantities of labelled data exist
for all languages at once. Such application scenarios, despite having
attracted less research attention than the few-shot and zero-shot
counterparts, are frequent in the context of multinational
organisations, such as the European Union or UNESCO, or multilingual
countries, such as India, South Africa, Singapore, and Canada, or
multinational companies (e.g., Amazon, Vodafone).
The aim of CLTC, in these latter cases, is to effectively exploit the
potential synergies among the different languages in order to allow
all languages to contribute to, and to benefit from, each other. Put
it another way, the \emph{raison d'être} of CLTC here becomes to
deploy classification systems that perform substantially better than
the trivial solution (the so-called \emph{naïve classifier})
consisting of $|\mathcal{L}|$ monolingual classifiers trained
independently of each other.




\subsection{Funnelling and Generalized Funnelling}

\noindent \citet{Esuli:2019dp} recently proposed \emph{Funnelling}
(\fun), an HTL method based on a two-tier classifier ensemble, and
applied it to CLTC. In \fun, the 1st-tier of the ensemble is composed
of $|\mathcal{L}|$ language-specific classifiers, one for each
language in $\mathcal{L}$. For each document $d$, one of these
classifiers (the one specific to the language of document $d$) returns
a vector of $|\mathcal{Y}|$ calibrated posterior probabilities, where
$\mathcal{Y}$ is the codeframe. Each such vector, irrespective of
which among \blue{the $\mathcal{L}$ classifiers} has generated it, is
then fed to a 2nd-tier ``meta-classifier'' which returns the final
label predictions.

The $|\mathcal{Y}|$-dimensional vector space to which the vectors of
posterior probabilities
belong, 
thus forms an ``interlingua'' among the $|\mathcal{L}|$ languages,
since all these vectors are homologous, independently of which among
the $|\mathcal{L}|$ classifiers have generated them. Another way of
saying it is that all vectors are \textit{aligned across languages},
i.e., the $i$-th dimension of the vector space has the same meaning in
every language (namely, the ``posterior'' probability that the
document belongs to class $y_i$). During training, the meta-classifier
can thus learn from all labelled documents, irrespectively of their
language. Given that the meta-classifier's prediction for each class
in $\mathcal{Y}$ depends on the posterior probabilities received in
input for all classes in $\mathcal{Y}$, the meta-classifier can
exploit class-class correlations, and this (among other things) gives
\fun\ an edge over CLTC systems in which these correlations cannot be
brought to bear.

\fun\ was originally conceived with the many-shot / few-shot setting
in mind; in such a setting, \fun\ proved superior to the naïve
classifier and to 6 state-of-the-art
baselines~\cite{Esuli:2019dp}. \citet{Esuli:2019dp} also sketched some
architectural modifications that allow \fun\ to be applied to the
zero-shot setting too.



In this paper we describe \emph{Generalized Funnelling} (\gfun), a
generalisation of \fun\ consisting of an HTL architecture in which
1st-tier components can be arbitrary \emph{view-generating functions}
(VGFs), i.e., language-dependent functions that each produce a
language-independent representation (``view'') of the
\blue{(monolingual)}  document. We
describe an instantiation of \gfun\ in which the meta-classifier
receives as input\blue{, for the same (monolingual) document,} a vector of calibrated posterior
probabilities (as in \fun) as well as other language-independent
vectorial representations, consisting of different types of document
embeddings. These additional vectors are aggregated (e.g., via
concatenation) with the original vectors of posterior probabilities,
and the result is a set of extended, language-aligned, heterogeneous
vectors, one for each \blue{monolingual} \label{monolingual5}
document.

The original \fun\ architecture is thus a particular instance of
\gfun, in which the 1st-tier is equipped with only one VGF.
The additional VGFs that characterize \gfun\
each enable the meta-classifier to gain access to information on types
of correlation in the data additional to the class-class correlations
captured by the meta-classifier. In particular, we investigate the
impact of \textit{word-class correlations} (as embodied in
\emph{Word-Class Embeddings} (WCEs)~\cite{Moreo:2021qq}),
\textit{word-word correlations} (as embodied in \emph{Multilingual
Unsupervised or Supervised Embeddings} (MUSEs) \cite{Conneau:2018bv}),
and \textit{correlations between contextualized words}
(as embodied in embeddings generated by \emph{multilingual
BERT}~\cite{Devlin:2018dl}). As we will show, \gfun\ natively caters
for both the many-shot/few-shot and the zero-shot settings; we carry
out extensive CLTC experiments in order to assess the performance of
\gfun\ in both cases. The results of these experiments show that
mining additional types of correlations in data does make a
difference, and that \gfun\ outperforms \fun\ as well as other CLTC
systems that have recently been proposed.

The rest of this article is structured as follows. In
Section~\ref{sec:themethod} we describe the \gfun\ framework, while in
Section~\ref{sec:viewgen} we formalize the concept of
``view-generating function'' and present several instances of it.
Section~\ref{sec:experiments} reports the experiments (for both the
many-shot and the zero-shot variants)\footnote{We do not explicitly
present experiments for the few-shot case since a few-shot system is
technically no different from a many-shot system.} that we have
performed on two large datasets for multilingual multilabel text
classification. In Section~\ref{sec:further} we move further and
discuss a more advanced, ``recurrent'' VGF that combines MUSEs and
WCEs in a more sophisticated way, and test it in additional
experiments. We review related work and methods in
Section~\ref{sec:relwork}. In Section~\ref{sec:conclusions} we
conclude by sketching avenues for further research. Our code that
implements \gfun\ is publicly
available.\footnote{\url{https://github.com/andreapdr/gFun}}

\section{Generalized Funnelling}\label{sec:themethod}

\noindent In this section, we first briefly summarise the original
\fun\ method, and then move on to present \gfun\ and related
concepts. 


\subsection{A brief introduction to Funnelling}
\label{sub:fun}

\noindent Funnelling, as described in~\cite{Esuli:2019dp}, comes in
two variants, called \ftat\ and \fkfcv. We here disregard \fkfcv\ and
only use \ftat, since in all the experiments reported
in~\cite{Esuli:2019dp} \ftat\ clearly outperformed \fkfcv;
see~\cite{Esuli:2019dp} if interested in a description of \fkfcv. For
ease of notation, we will simply use \fun\ to refer to \ftat.

\begin{figure}[t]
  \centering
  \includegraphics[width=1\textwidth]{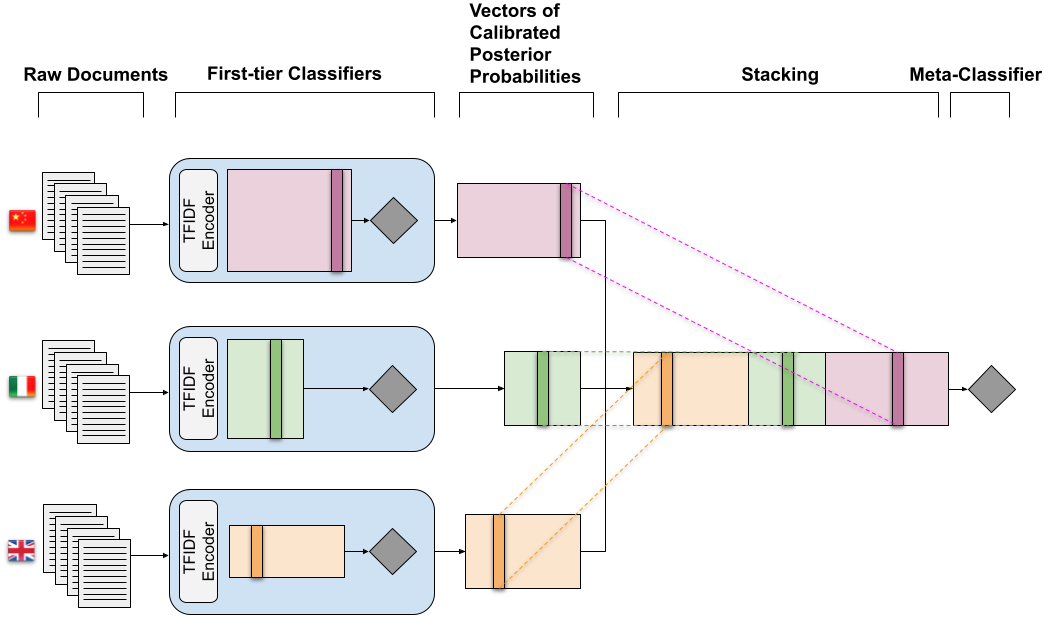}
  \caption{The \fun\ architecture, exemplified with $|\mathcal{L}|$=3
  languages (Chinese, Italian, English). Note that the different
  term-document matrices in the 1st-tier may contain different numbers
  of documents and/or different numbers of terms. The three grey
  diamonds on the left represent calibrated classifiers that map the
  original vectors (e.g., TFIDF vectors) into
  $|\mathcal{Y}|$-dimensional spaces. The resulting vectors are thus
  aligned and can all be used for training the meta-classifier, which
  is represented by the grey diamond on the right.}
  \label{fig:vanilla_fun}
\end{figure}

In \fun\ (see Figure~\ref{fig:vanilla_fun}), in order to train a
classifier ensemble, 1st-tier language-specific classifiers
$h^{1}_{1},..., h^{1}_{|\mathcal{L}|}$ (with superscript $1$
indicating the $1$st tier) are trained from their corresponding
language-specific training sets
$\mathrm{Tr}_{1},...,\mathrm{Tr}_{|\mathcal{L}|}$. Training documents
$d\in \mathrm{Tr}_{i}$ may be represented by means of any desired
vectorial representation $\phi^{1}_{i}(d)=\mathbf{d}$, such as, e.g.,
$\tfidf$-weighted bag-of-words, or character $n$-grams; in principle,
different styles of vectorial representation can be used for the
different 1st-tier classifiers, if desired. The classifiers may be
trained by any learner, provided the resulting classifier returns, for
each language $\lambda_{i}$, document $d$, and class $y_{j}$, a
confidence score
$h^{1}_{i}(\mathbf{d},y_{j})\in\mathbb{R}$; in principle, different
learners can be used for the different 1st-tier classifiers, if
desired.

Each 1st-tier classifier $h^{1}_{i}$ is then applied to each training
document $d\in \mathrm{Tr}_{i}$, thus generating a vector
\begin{align}
  \begin{split}
    \label{eq:scores}
    S(d) = & \ (h^{1}_{i}(\mathbf{d},y_{1}), ...,
    h^{1}_{i}(\mathbf{d},y_{|\mathcal{Y}|}))
  \end{split}
\end{align}
\noindent of confidence scores for each $d\in
\mathrm{Tr}_{i}$. (\label{line:funtat}Incidentally, this is the phase
in which \ftat\ and \fkfcv\ differ, since \fkfcv\ uses instead a
$k$-fold cross-validation process to classify the training documents.)

The next step consists of computing (via a chosen probability
calibration method) language- and class-specific calibration functions
$f_{ij}$ that map confidence scores $h^{1}_{i}(\mathbf{d},y_{j})$ into
calibrated posterior probabilities
$\Pr(y_{j}|\mathbf{d})$.\footnote{\label{foot:calibration}\blue{The
reason why we need calibration is that the confidence scores obtained
from different classifiers are not comparable; the calibration process
serves the purpose of mapping these confidence scores into entities
(the calibrated posterior probabilities) that are indeed comparable
even if originating from different classifiers.}}

\fun\ then applies $f_{ij}$ to each confidence score
and obtains a vector of calibrated posterior probabilities
\begin{align}
  \begin{split}
    \label{eq:posteriors}
    \phi^{2}(d) = & \ (f_{i1}(h^{1}_{i}(\mathbf{d},y_{1})),...,
    f_{i|\mathcal{Y}|}(h^{1}_{i}(\mathbf{d},y_{|\mathcal{Y}|}))) \\ =
    & \ (\Pr(y_{1}|\mathbf{d}),..., \Pr(y_{|\mathcal{Y}|}|\mathbf{d}))
  \end{split}
\end{align}
\noindent Note that the $i$ index for language $\lambda_{i}$ has
disappeared, since calibrated posterior probabilities are comparable
across different classifiers, which means that we can use a shared,
language-independent space of vectors of calibrated posterior
probabilities.

At this point, the 2nd-tier, language-independent ``meta''-classifier
$h^{2}$ can be trained from all training documents
$d\in\bigcup_{i=1}^{|\mathcal{L}|}\mathrm{Tr}_{i}$, where document $d$
is represented by its $\phi^{2}(d)$ vector.
This concludes the training phase.

In order to apply the trained ensemble to a test document
$d\in \mathrm{Te}_{i}$ from language $\lambda_{i}$, \fun\
applies classifier $h^1_{i}$ to $\phi^{1}_{i}(d)=\mathbf{d}$ and
converts the resulting vector $S(d)$ of confidence scores into a
vector $\phi^{2}(d)$ of calibrated posterior probabilities. \fun\ then
feeds this latter to the meta-classifier $h^{2}$, which returns (in
the case of multilabel classification) a vector of binary labels
representing the predictions of the meta-classifier.


\subsection{Introducing heterogeneous correlations through Generalized
Funnelling} \label{sub:gfun}

\noindent As explained in~\cite{Esuli:2019dp}, the reasons why \fun\
outperforms the naïve monolingual baseline consisting of
$|\mathcal{L}|$ independently trained, language-specific classifiers,
are essentially two. The first is that \fun\ learns from heterogeneous
data; i.e., while in the naïve monolingual baseline each classifier is
trained only on $|\mathrm{Tr}_{i}|$ labelled examples, the
meta-classifier in \fun\ is trained on all the
$\bigcup_{i=1}^{|\mathcal{L}|}|\mathrm{Tr}_{i}|$ labelled examples.
Put it another way, in \fun\ all training examples contribute to
classifying all unlabelled examples, irrespective of the languages of
the former and of the latter.
The second is that the meta-classifier leverages \emph{class-class
correlations}, i.e., it learns to exploit the stochastic dependencies
between classes typical of multiclass settings. In fact, for an
unlabelled document $d$ the meta-classifier receives $|\mathcal{Y}|$
inputs from the 1st-tier classifier which has classified $d$, and
returns $|\mathcal{Y}|$ confidence scores, which means that the input
for class $y'$ has a potential impact on the output for class $y''$,
for every $y'$ and $y''$.


In \fun, the key step in allowing the meta-classifier to leverage the
different language-specific training sets consists of mapping all the
documents onto a space shared among all
languages. 
This is made possible by the fact that the 1st-tier classifiers all
return vectors of calibrated posterior probabilities.
These vectors are homologous (since the codeframe is the same for all
languages), and are also comparable (because the posterior
probabilities are calibrated), which means that we can have all
vectors share the same vector space irrespectively of the language of
provenance.

In \gfun, we generalize this mapping by allowing a set $\Psi$ of
\emph{view-generating functions} (VGFs) to define this shared vector
space. VGFs are language-dependent functions that map
\blue{(monolingual)} \label{line:monolingual3} documents into
language-independent vectorial representations (that we here call
\emph{views}) aligned across languages. Since each view is aligned
across languages, it is easy to aggregate (e.g., by concatenation) the
different views \blue{of the same monolingual document}
\label{line:monolingual6} into a single representation that is also
aligned across languages, and which can be thus fed to the
meta-classifier.

Different VGFs are meant to encode different types of information so
that they can all be brought to bear on the training process. In the
present paper we will experiment with extending \fun\ by allowing
views consisting of different types of document embeddings, each
capturing a different type of correlation within the data.

The procedures for training and testing cross-lingual classifiers via
\gfun\ are described in Algorithm~\ref{alg:gfun:train} and
Algorithm~\ref{alg:gfun:test}, respectively.
The first step of the training phase is the optimisation of the
parameters (if any) of the VGFs $\psi_{k}\in\Psi$
(Algorithm~\ref{alg:gfun:train} -- Line~\ref{line:fit}), which is
carried out independently for each language and for each VGF. A VGF
$\psi_k$ produces representations that are aligned across all
languages, which means that vectors coming from different languages
can be ``stacked'' (i.e., placed in the same set) to define the view
$V_k$ (Algorithm~\ref{alg:gfun:train} --
Line~\ref{line:vstack:train}), which corresponds to the $\psi_k$
portion of the entire (now language-independent) training set of the
meta-classifier. Note that the vectors in a given view need not be
probabilities; we only assume that they are homologous and comparable
across languages. The aggregation function ($\mathit{aggfunc}$)
implements a policy for aggregating the different views for them to be
input to the meta-classifier; it is thus used both during training
(Algorithm~\ref{alg:gfun:train} -- Line~\ref{line:aggfunc:train}) and
during test (Algorithm~\ref{alg:gfun:test} --
Line~\ref{line:aggfunc:test}). In case the aggregation function needs
to learn some parameters, those are estimated during training
(Algorithm~\ref{alg:gfun:train} -- Line~\ref{line:fit:aggfunc}).

Finally, note that both the training phase and the test phase are
highly parallelisable, since the (training and/or testing) data for
language $\lambda'$ can be processed independently of the analogous
data for language $\lambda''$, and since each view within a given
language can be generated independently of the other views for the
same language.
 
\begin{algorithm}[t]
  \LinesNumbered \SetNoFillComment
  \begin{footnotesize}
    \SetKwInOut{Input}{Input} \SetKwInOut{Output}{Output} \Input{
    \textbullet\ Sets
    $\{\mathrm{Tr}_{1},...,\mathrm{Tr}_{|\mathcal{L}|}\}$ of training
    documents written in languages
    $\mathcal{L}=\{\lambda_{1},...,\lambda_{|\mathcal{L}|}\}$, all
    labelled according to
    $\mathcal{Y}=\{y_{1},..., y_{|\mathcal{Y}|}\}$; \\
    \hspace{0.25em}\textbullet\ Set $\Psi=\{\psi_{1},...,\psi_{|\Psi|}\}$ of VGFs; \\
    }\Output{\textbullet\ VGF parameters
    $\Theta=\{\theta_{ik}\}, 1\leq i\leq |\mathcal{L}|,
    1\leq k\leq |\Psi|$ ; \\
    \hspace{0.25em}\textbullet\ Parameters of the aggregation function $\Lambda$ \\
    \hspace{0.25em}\textbullet\ Meta-classifer $h^{2}$ }\BlankLine

    \For{$\psi_{k}\in\Psi$}{ \tcc{Learn the parameters of the $k$th
    VGF for each language $\lambda_{i}$}
    \For{$\lambda_{i}\in \mathcal{L}$}{
    $\theta_{ik}\leftarrow \mathit{fit}(\psi_{k}, \mathrm{Tr}_{i})$
    ; \label{line:fit} }

    \tcc{Stack all language views produced by $\psi_{k}$}
    $\mathrm{V}_{k}\leftarrow
    \mathrm{vstack}(\psi_k(\mathrm{Tr}_{1},\theta_{1k}), \ldots,
    \psi_{k}(\mathrm{Tr}_{|\mathcal{L}|},\theta_{|\mathcal{L}|k}))$
    ; \label{line:vstack:train} }

    \tcc{Learn the parameters (if any) of the aggregation function}
    $\Lambda\leftarrow \mathit{fit}(\mathit{aggfunc},\ldots)$
    ; \label{line:fit:aggfunc}
 
    \tcc{Combine all training sets by aggregating the
    language-independent views}
    $\mathrm{Tr}'\leftarrow \mathit{aggfunc}(V_{1}, \ldots ,
    V_{|\Psi|}, \Lambda)$ ; \label{line:aggfunc:train}

    \BlankLine
    Train meta-classifier $h^{2}$ from all vectors in $\mathrm{Tr}'$ ; \\
    $\Theta\leftarrow\{\theta_{ik}\}, 1\leq i\leq |\mathcal{L}|,
    1\leq k\leq |\Psi|$ ; \\
    \Return{$\Lambda$, $\Theta$, $h^{2}$} \BlankLine

    \caption{Generalized Funnelling for CLTC, training phase.}
    \label{alg:gfun:train}
  \end{footnotesize}
\end{algorithm}

\begin{algorithm}[t]
  \LinesNumbered \SetNoFillComment
  \begin{footnotesize}
    \SetKwInOut{Input}{Input} \SetKwInOut{Output}{Output} \Input{
    \textbullet\ Sets $\{\mathrm{Te}_{1},...,\mathrm{Te}_{|\mathcal{L}|}\}$ of unlabelled documents written in languages $\mathcal{L}=\{\lambda_{1},...,\lambda_{|\mathcal{L}|}\}$, all to be labelled according to $\mathcal{Y}=\{y_{1},..., y_{|\mathcal{Y}|}\}$; \\
    \hspace{0.25em}\textbullet\ Set
    $\Psi=\{\psi_{1},...,\psi_{|\Psi|}\}$ of VGFs with parameters
    $\Theta=\{\theta_{ik}\}, 1\leq i\leq |\mathcal{L}|,
    1\leq k\leq |\Psi|$ ; \\
    \hspace{0.25em}\textbullet\ Parameters
    $\Lambda$ of the aggregation function ; \\
    \hspace{0.25em}\textbullet\ meta-classifier $h^{2}$ ; \\
    }\Output{\textbullet\ Labels for all documents in $\{\mathrm{Te}_{1},...,\mathrm{Te}_{|\mathcal{L}|}\}$ ; \\
    }\BlankLine
    \For{$\lambda_{i}\in \mathcal{L}$}{

    \tcc{Aggregate the views produced by all VGFs}

    $\mathrm{Te}_{i}'\leftarrow \mathit{aggfunc}(\psi_1(\mathrm{Te}_{i},\theta_{i1}), \ldots, \psi_{|\Psi|}(\mathrm{Te}_{i},\theta_{i|\Psi|}), \Lambda)$ ; \label{line:aggfunc:test} \\
    \tcc{Use the meta-classifier $h^{2}$ to predict labels $L_{i}$ for
    all documents in $\mathrm{Te}_{i}'$}
    $L_{i} \leftarrow h^{2}(\mathrm{Te}_{i}')$ }
    \Return{$\{L_{1},\ldots,L_{|\mathcal{L}|}\}$} \BlankLine

    \caption{Generalized Funnelling for CLTC, testing phase.
    }\label{alg:gfun:test}
  \end{footnotesize}
\end{algorithm}

Note that the original formulation of \fun\ (Section~\ref{sub:fun})
thus reduces to an instance of \gfun\ in which there is a single VGF
(one that converts documents into calibrated posterior probabilities)
and the aggregation function is simply the identity function. In this
case, the fit of the VGF (Algorithm~\ref{alg:gfun:train} --
Line~\ref{line:fit}) comes down to computing weighted (e.g., via
TFIDF) vectorial representations of the training documents, training
the 1st-tier classifiers, and calibrating them. Examples of the
parameters obtained as a result of the fitting process include the
choice of vocabulary, the IDF scores, the parameters of the separating
hyperplane, and those of the calibration function.
During the test phase, invoking the VGF (Algorithm~\ref{alg:gfun:test}
-- Line~\ref{line:aggfunc:test}) amounts to computing the weighted
vectorial representations and the $\phi^2(d)$ representations
(Equation~\ref{eq:posteriors}) of the test documents, using the
classifiers and meta-classifier generated during the training
phase.

In what follows we describe the VGFs that we have investigated in
order to introduce into \gfun\ sources of information additional to
the ones that are used in \fun. In particular, we describe in detail
each such VGF in Sections \ref{sec:vg_vanillafun}-\ref{sec:vg_mbert},
we discuss aggregation policies in Section~\ref{sec:aggfun}, and we
analyse a few additional modifications concerning data normalisation
(Section \ref{sec:normalisation}) that we have introduced into \gfun\
and that, although subtle, bring about a substantial improvement in
the effectiveness of the method.


\section{View-generating functions}\label{sec:viewgen}

\noindent In this section we describe the VGFs that we have
investigated throughout this research, by also briefly explaining
related concepts and works from which they stem.


As already stated, the main idea behind our instantiation of \gfun\ is
to learn from heterogeneous information about different kinds of
correlations in the data. While the main ingredients of the text
classification task are words, documents, and classes, the key to
approach the CLTC setting lies in the ability to model them
consistently across all languages. \label{line:signaljustification} We
envision ways for bringing to bear the following stochastic
correlations among these elements:

\begin{enumerate}
\item Correlations between different classes: understanding how
  classes are related to each other in some languages may bring about
  additional knowledge useful for classifying documents in other
  languages. These correlations are specific to the particular
  codeframe used, and are obviously present only in
  \blue{\label{line:multilabel}multilabel} scenarios. They can be used
  (in our case: by the meta-classifier) in order to refine an initial
  classification (in our case: by the 1st-tier classifiers), since
  they are based on the relationships between posterior probabilities
  / labels assigned to documents.

 

\item Correlations between different words: by virtue of the
  ``distributional hypothesis'' (see~\citep{Sahlgren:2006yq}), words
  are often modelled in accordance to how they are distributed in
  corpora of text with respect to other words. Distributed
  \blue{representations} of words encode the relationships between
  words and other words; when properly aligned across languages, they
  represent an important help for bringing lexical semantics to bear
  on multilingual text analysis processes, thus helping to bridge the
  gap among language-specific sources of labelled information.

\item Correlations between words and classes: profiling words in terms
  of how they are distributed across the classes in a language is a
  direct way of devising cross-lingual word embeddings (since
  translation-equivalent words are expected to exhibit similar
  class-conditional distributions), which is compliant with the
  distributional hypothesis (since semantically similar words are
  expected to be distributed similarly across classes).

\item Correlations between contextualized words: the meaning of a word
  occurrence is dependent on the specific context in which the word
  occurrence is found. Current language models are well aware of this
  fact, and try to generate contextualized representations of words,
  which can in turn be used straightforwardly in order to obtain
  contextualized representations for entire documents. Language models
  trained on multilingual data are known to produce distributed
  representations that are coherent across the languages they have
  been trained on.

\end{enumerate}

\noindent We recall from Section~\ref{sub:fun} that class-class
correlations are exploited in the 2nd-tier of \fun. We model the other
types of correlations mentioned above via dedicated VGFs. \blue{We
investigate instantiations of the aforementioned correlations by means
of independently motivated modular VGFs. Here we provide a brief overview of each
them.}\label{line:RaC8}

\begin{itemize}

\item \textit{the Posteriors VGF}:
  it maps documents into the space defined by calibrated posterior
  probabilities. This is, aside from the modifications discussed in
  Section~\ref{sec:normalisation}, equivalent to the 1st-tier of the
  original \fun, but we discuss it in detail in
  Section~\ref{sec:vg_vanillafun}.

\item \textit{the MUSEs VGF} (encoding correlations between different
  words): it uses the \blue{(supervised version
  of)}\label{line:MUSEs} Multilingual Unsupervised or Supervised
  Embeddings (MUSEs) made available by the authors
  of~\cite{Conneau:2018bv}. MUSEs have been trained on
  Wikipedia\footnote{https://dumps.wikimedia.org/} in 30 languages and
  have later been aligned using bilingual dictionaries and iterative
  Procrustes alignment (see Section~\ref{sec:vg_muse} and
  \citep{Conneau:2018bv}).

\item \textit{the WCEs VGF} (encoding correlations between words and
  classes): it uses Word-Class Embeddings (WCEs)~\cite{Moreo:2021qq},
  a form of supervised word embeddings based on the class-conditional
  distributions observed in the training set (see
  Section~\ref{sec:vg_wce}).

 

\item \textit{the BERT VGF} (encoding correlations between different
  contextualized words): it uses the contextualized word embeddings
  generated by multilingual BERT~\cite{Devlin:2019xx}, a deep
  pretrained language model based on the transformer architecture (see
  Section~\ref{sec:vg_mbert}).


\end{itemize}


\noindent In the following sections we present each VGF in detail.
 

\subsection{The Posteriors VGF}\label{sec:vg_vanillafun}

\noindent This VGF coincides with the 1st-tier of \fun, but we briefly
explain it here for the sake of completeness.

Here the idea is to leverage the fact that the classification scheme
is common to all languages, in order to define a vector space that is
aligned across all languages. Documents, regardless of the language
they are written in, can be redefined with respect to their relations
to the classes in the codeframe. \blue{\label{line:geometrically}Using
a geometric metaphor}, the relation between a document and a class can
be defined in terms of the distance between the document and the
surface that separates the class from its complement. In other words,
while the language-specific vector spaces where the original document
vectors lie are not aligned (e.g., they can be characterized by
different numbers of dimensions, and the dimensions for one language
bear no relations to the dimensions for another language), one can
profile each document via a new vector consisting of the distances to
the separating surfaces relative to the various classes. By using the
binary classifiers as ``pivots'' \citep{Ando:2005fr}, documents end up
being represented in a shared space, in which the number of dimensions
are the same for all languages (since the classes are assumed to be
the same for all languages), and the vector values for each dimension
are comparable across languages once the distances to the
classification surfaces are properly normalized (which is achieved by
the calibration process).

Note that this procedure is, in principle, independent of the
characteristics of any particular vector space and learning device
used across languages, both of which can be different across the
languages.\footnote{The vector spaces can indeed be completely
different from one language to another. For example, one could define
a bag of TFIDF-weighted bigrams for English, a bag of BM25-weighted
unigrams for French, and an SVD-decomposed space for Spanish. Note
also that the learning algorithms can be different as well; one may
use, say, SVMs for English, logistic regression for French, and
AdaBoost for Spanish. As long as the decision scores provided by each
classifier are turned into calibrated posterior probabilities, the
language-specific representations can be recast into
language-independent, comparable representations.}

For ease of comparability with the results reported
by~\citet{Esuli:2019dp}, in this paper we will follow these authors
and encode (for all languages in $\mathcal{L}$) documents as
bag-of-words vectors weighted via $\tfidf$, which is computed as
\begin{align}\label{eq:tfidf}
  \tfidf(w_{k},\textbf{x}_{j}) = \tf(w_{k},\textbf{x}_{j})\cdot \log \frac{|\mathrm{Tr}|}{\text{\#}_{\mathrm{Tr}}(w_{k})}
\end{align}
\noindent where $\text{\#}_{\mathrm{Tr}}(w_{k})$ is the number of
documents in $\mathrm{Tr}$ in which word $w_{k}$ occurs at least once
and
\begin{align}
  \tf(w_{k},\textbf{x}_{j}) = \begin{cases} 1+\log \text{\#}(w_{k},\textbf{x}_{j}) & \mbox{if } \text{\#}(w_{k},\textbf{x}_{j}) > 0 \\ 0 & \text{otherwise} \end{cases}
                                                                                                                                           \label{eq:tf}
\end{align}
\noindent where $\text{\#}(w_{k},\textbf{x}_{j})$ stands for the
number of times $w_{k}$ appears in document $\textbf{x}_{j}$. Weights
are then normalized via cosine normalisation, as
\begin{equation}
  w(w_{k},\textbf{x}_{j})=\frac{\tfidf(w_{k},\textbf{x}_{j})}{\sqrt{\sum_{w'\in d_{j}} \tfidf(w'_{k},\textbf{x}_{j})^2}}
  \label{eq:tfidfnorm}
\end{equation}


\noindent For the very same reasons we also follow~\cite{Esuli:2019dp}
in adopting (for all languages in $\mathcal{L}$) Support Vector
Machines (SVMs) as the learning algorithm, and ``Platt
calibration''~\cite{Platt:2000fk} as the probability calibration
function.


\subsection{The MUSEs VGF}\label{sec:vg_muse}

\noindent In CLTL, the need to transfer lexical knowledge across
languages has given rise to cross-lingual representations of words in
a joint space of embeddings.
In our research, in order to encode word-word correlations across
different languages we derive document embeddings from \blue{(the
supervised version of)}\label{line:MUSEs2} \textit{Multilingual
Unsupervised or Supervised Embeddings} (MUSEs)~\cite{Conneau:2018bv}.
MUSEs are word embeddings generated via a method for aligning
unsupervised (originally monolingual) word embeddings in a shared
vector space,
similar to the method described in~\cite{Mikolov:2013no}. The
alignment is obtained via a linear mapping (i.e., a rotation
matrix) 
learned by an adversarial training process in which a \emph{generator}
(in charge of mapping the source embeddings onto the target space) is
trained to fool a \emph{discriminator} from distinguishing the
language of provenance of the embeddings, i.e., from discerning if the
embeddings it receives as input originate from the target language or
are instead the product of a transformation of embeddings originated
from the source language. The mapping 
is then further refined using a technique called ``Procrustes
alignment''. The qualification ``Unsupervised or Supervised" refers to
the fact that the method can operate with or without a dictionary of
parallel seed words; \label{line:MUSEs3}\blue{we use the
embeddings generated in supervised fashion.}

We use the MUSEs that~\citet{Conneau:2018bv} make publicly
available\footnote{\url{https://github.com/facebookresearch/MUSE}},
and that consist of 300-dimensional multilingual word embeddings
trained on Wikipedia using fastText. To date, the embeddings have been
aligned for 30 languages with the aid of bilingual dictionaries.

Fitting the VGF for MUSEs consists of first allocating in memory the
pre-trained MUSE matrices $\mathbf{M}_{i}$
\blue{$\in \mathbb{R}^{(v_i \times 300)}$ (where $v_i$ is the
vocabulary size for the $i$-th language)}, made available
by~\citet{Conneau:2018bv}, for each language $\lambda_{i}$ involved,
and then generating document embeddings for all training documents as
weighted averages of the words in the document. As the weighting
function, we use TFIDF (Equation \ref{eq:tfidf}). This computation
reduces to performing the projection
$\mathbf{X}_{i}\cdot\mathbf{M}_{i}$, where the matrix $\mathbf{X}_{i}$
\blue{$\in \mathbb{R}^{(|\mathrm{Tr_{i}}| \times v_i)}$} consists of
the $\tfidf$-weighted vectors that represent the training documents
(for this we can reuse the matrices $\mathbf{X}_{i}$ computed by the
Posteriors VGF, since they are identical to the ones needed here). The
process of generating the views of test documents via this VGF is also
obtained via a projection $\mathbf{X}_{i}\cdot\mathbf{M}_{i}$, where
in this case the $\mathbf{X}_{i}$ matrix consists of the
$\tfidf$-weighted vectors that represent the \emph{test}
documents. 



\subsection{The WCEs VGF}\label{sec:vg_wce}

\noindent In order to encode word-class correlations we derive
document embeddings from \textit{Word-Class Embeddings}
(WCEs~\cite{Moreo:2021qq}).
WCEs are supervised embeddings meant to extend (e.g., by
concatenation) other unsupervised pre-trained word embeddings (e.g.,
those produced by means of \texttt{word2vec}, or \texttt{GloVe}, or
any other technique) in order to
inject task-specific word meaning in multiclass text classification.
%
The WCE for word $w$ is defined as
\begin{align}
  E(w) = \varphi(\eta(w,y_{1}),. . .,\eta(w,y_{|\mathcal{Y}|}))
\end{align}
\noindent where $\eta$ is a real-valued function that quantifies the
correlation between word $w$ and class $y_{j}$ as observed in the
training set, and where $\varphi$ is any dimensionality reduction
function. Here, as the $\eta$ function we adopt the normalized dot
product, as proposed in~\cite{Moreo:2021qq}, whose computation is very
efficient; as $\varphi$ we use the identity function, which means that
our WCEs are $|\mathcal{Y}|$-dimensional vectors.

So far, WCEs have been tested exclusively in monolingual
settings. However, WCEs are \emph{naturally aligned across languages},
since WCEs have one dimension for each $y\in \mathcal{Y}$, which is
the same for all languages $\lambda_{i}\in \mathcal{L}$. Document
embeddings relying on WCEs thus display similar characteristics
irrespective of the language in which the document is written in. In
fact, given a set of documents classified according to a common
codeframe, WCEs reflect the intuition that words that are semantically
similar across languages (i.e., are translations of each other) tend
to exhibit similar correlations to the classes in the codeframe. This
is, to the best of our knowledge, the first application of WCEs to a
multilingual setting.

\begin{figure}[t] \centering
  \includegraphics[width=1\textwidth]{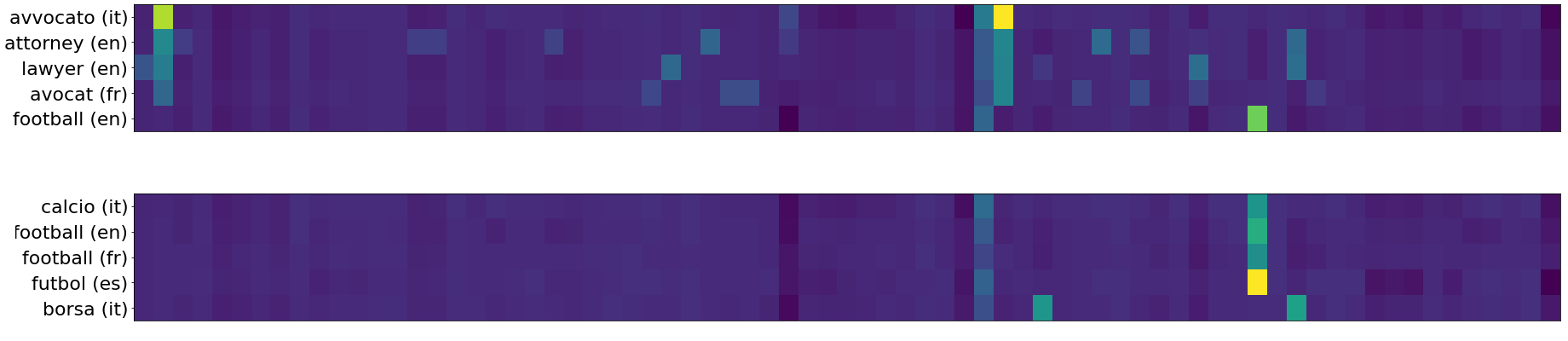}
  \caption{Heatmaps displaying five WCEs each, where each cell
  indicates the correlation between a word (row) and a class (column),
  as from the RCV1/RCV2 dataset. Yellow indicates a high value of
  correlation while blue indicates a low such value. Words
  ``avvocato'' and ``avocat'' are Italian and French translations,
  resp., of the English word ``lawyer''; words ``calcio'' and
  ``futbol'' are Italian and Spanish translations, resp., of the
  English word ``football''; Italian word ``borsa'' instead means
  ``bag''. }
  \label{fig:WCE_examples}
\end{figure}

The intuition behind this idea is illustrated by the two examples in
Figure~\ref{fig:WCE_examples}, where two heatmaps display the
correlation values of five WCEs each. Each of the two heatmaps
illustrates the distribution patterns of four terms that are either
semantically related or translation equivalents of each other (first
four rows in each subfigure), and
of a fifth term semantically unrelated to the previous four (last row
in each subfigure).
Note that not only semantically related terms in a language get
similar representations (as is the case of ``attorney'' and ``lawyer''
in English), but also translation-equivalent terms do so (e.g.,
``avvocato'' in Italian and ``avocat'' in French).

The VGF for WCEs is similar to that for MUSEs, but for the fact that
in this case the matrix containing the word embeddings needs to be
obtained from our training data, and is not pretrained on external
data. More specifically, fitting the VGF for WCEs comes down to first
computing, for each language $\lambda_{i}\in \mathcal{L}$, the
language-specific WCE matrix $\mathbf{W}_{i}$ according to the process
outlined in~\cite{Moreo:2021qq},
and then projecting the TFIDF-weighted matrix $\mathbf{X}_{i}$
obtained from $\mathrm{Tr}_{i}$,
as $\mathbf{X}_{i}\cdot\mathbf{W}_{i}$. (Here too, we use the TFIDF
variant of Equation~\ref{eq:tfidf}.) During the testing phase, we
simply perform the same projection $\mathbf{X}_{i}\cdot\mathbf{W}_{i}$
as above, where $\mathbf{X}_{i}$ now represents the weighted matrix
obtained from the test set.

\blue{Although alternative ways of exploiting word-class correlations
have been proposed in the literature, we adopted WCEs because of their
higher simplicity with respect to other methods. For example, the GILE
system \cite{gile} uses label descriptions in order to compute a model
of compatibility between a document embedding and a label embedding;
differently from the latter work, in our problem setting we do not
assume to have access to textual descriptions of the semantics of the
labels. The LEAM model \cite{leam}, instead, defines a word-class
attention mechanism and can work with or without label descriptions
(though the former mode is considered preferable), but has never been
tested in multilingual contexts; preliminary experiments we have
carried out by replacing the GloVe embeddings originally used in LEAM
with MUSE embeddings, have not produced competitive results.}


\subsection{The BERT VGF}\label{sec:vg_mbert}


\noindent BERT~\cite{Devlin:2019xx} is a bidirectional language model
based on the transformer architecture~\cite{Vaswani:2017hu} trained on
a \textit{masked language modelling} objective and \textit{next
sentence prediction} task. The transformer architecture has been
initially proposed for the task of sequence transduction relying
solely on the attention mechanism, and thus discarding the usual
recurrent components deployed in \blue{encoder-decoder} architectures.
BERT's transformer blocks contain two sub-layers. The first is a
multi-head self-attention mechanism, and the second is a simple,
position-wise fully connected feed-forward network. Differently from
other architectures~\citep{Peters:2018zt}, BERT's attention is set to
attend to all the input tokens (i.e., it deploys bidirectional
self-attention), thus making it well-suited for sentence-level
tasks. Originally, the BERT architecture was trained by
\citet{Devlin:2019xx} on a monolingual corpus composed of the
BookCorpus and English Wikipedia (for a total of roughly 3,300M
words). Recently, a multilingual version, called
mBERT~\cite{Devlin:2018dl}, has been released. The model is no
different from the standard BERT model; however, mBERT has been
trained on concatenated documents gathered from Wikipedia in 104
different languages. Its multilingual capabilities
emerge from the exposure to different languages during this massive
training phase.

In this research, we explore mBERT as a VGF for \gfun.
At training time, this VGF is first equipped with a fully-connected
output layer, so that BERT can be trained end-to-end using binary
cross-entropy as the loss function. Nevertheless, as its output (i.e.,
the one that is eventually fed to the meta-classifier also at testing
time) we use the hidden state associated with the document embedding
(i.e., the \texttt{[CLS] token)} at its last layer.




\subsection{Policies for aggregating VGFs}\label{sec:aggfun}

\noindent The different views of the same document that are
independently generated by the different VGFs need to be somehow
merged together before being fed to the meta-classifier. This is
undertaken by operators that we call \textit{aggregation functions}.
We explore two different policies for view aggregation: concatenation
and averaging.

\emph{Concatenation} simply consists of juxtaposing, for a given
document, the different views of this document, thus resulting in a
vector whose dimensionality is the sum of the dimensionalities of the
contributing views. This policy is the more straightforward one, and
one that does not impose any constraint on the dimensionality of the
individual views as generated from different VGFs.

\emph{Averaging} consists instead of computing, for a given document,
a vector which is the average of the different views for this
document. In order for it to be possible, though, this policy requires
that the views (i) all have the same dimensionality, and (ii) are
aligned among each other, i.e., that the $i$-th dimension of the
vector has the same meaning in every view. This is obviously not the
case with the views produced by the VGFs we have described up to now.
In order to solve this problem, we learn additional mappings onto the
space of class-conditional posterior probabilities, i.e., for each VGF
(other than the Posteriors VGF of Section~\ref{sec:vg_vanillafun},
which already returns vectors of $|\mathcal{Y}|$ calibrated posterior
probabilities) we train a classifier that maps the view of a document
into a vector of $|\mathcal{Y}|$ calibrated posterior
probabilities. The net result is that each document $d$ is represented
by $m$ vectors of $|\mathcal{Y}|$ calibrated posterior probabilities
(where $m$ is the number of VGFs in our system). These $m$ vectors can
be averaged, and the resulting average vector can be fed to the
meta-classifier as the only representation of document $d$. The way we
learn the above mappings is the same used in \fun; this also brings
about uniformity between the vectors of posterior probabilities
generated by the Posteriors VGF and the ones generated by the other
VGFs. Note that in this case, though, the classifier for VGF $\psi_k$
is trained on the views produced by $\psi_k$ for \textit{all} training
documents, irrespectively of their language of provenance; in other
words, for performing these mappings we just train $(m-1)$ (and not
$(m-1)\times |\mathcal{L}|$) classifiers, one for each VGF other than
the Posteriors VGF.




\begin{figure}[th]
  \centering
  \includegraphics[width=1\textwidth]{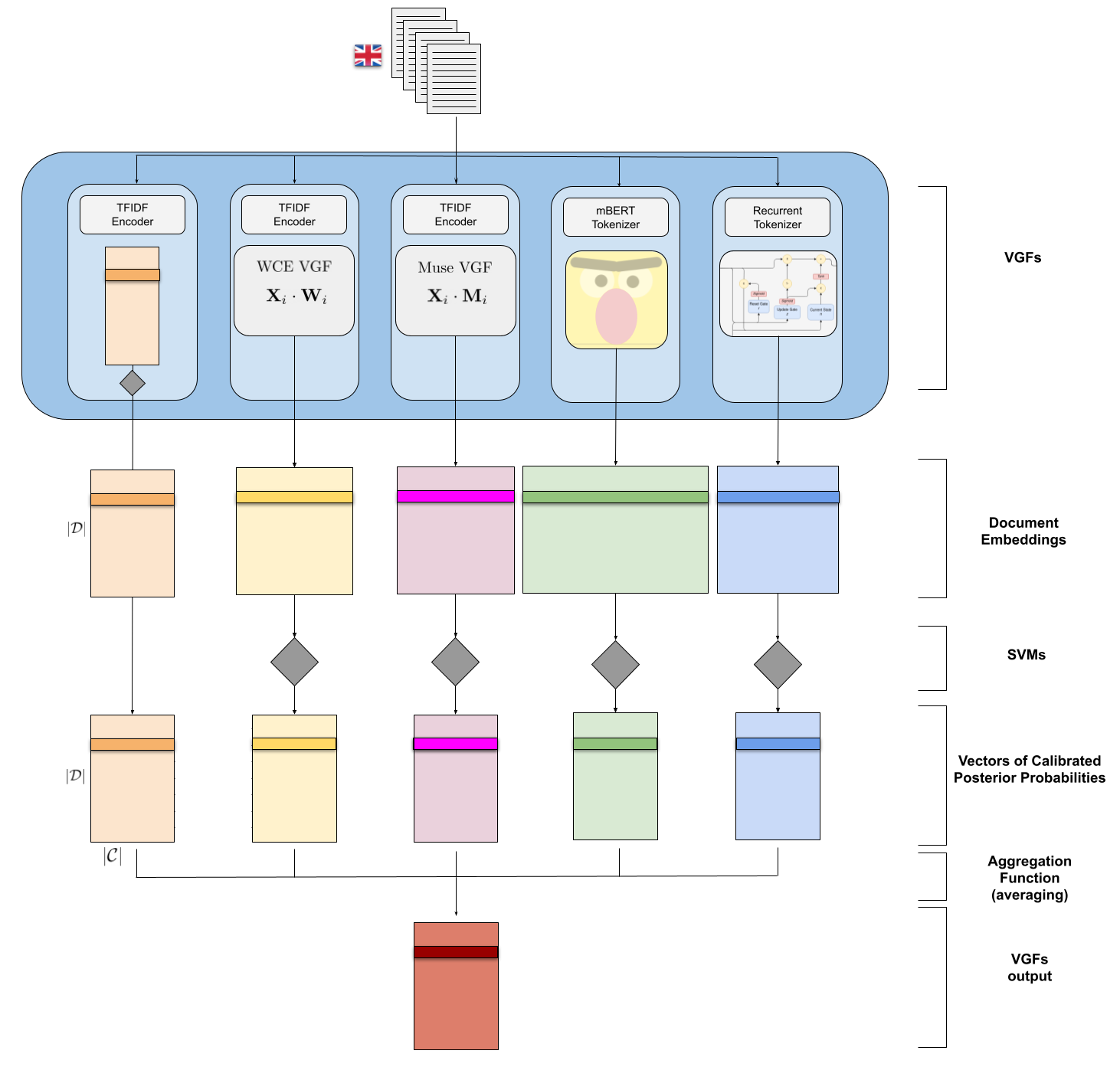}
  \caption{The averaging policy for view aggregation: the views are
  recast in terms of vectors of calibrated posterior probabilities
  before being
  averaged. 
  Note that the resulting vectors lie in the same vector space. For
  ease of visualisation, only one language (English) is shown.}
  \label{fig:VG_general}
\end{figure}

Each of these two aggregation policies has different pros and cons.

The main advantage of concatenation is that it is very simple to
implement. However, it suffers from the fact that the number of
dimensions in the resulting dense vector space is high, thus leading
to a higher computational cost for the meta-classifier.
Above all, since the number of dimensions that the different views
contribute is not always the same, this space (and the decisions of
the meta-classifier) can be eventually dominated by the VGFs
characterized by the largest number of dimensions.

The averaging policy (Figure \ref{fig:VG_general}), on the other hand,
scales well with the number of VGFs, but requires learning additional
mappings aimed at homogenising the different views into a unified
representation that allows averaging them. Despite the additional
cost, the averaging policy has one appealing characteristic, i.e.,
\emph{the 1st-tier is allowed to operate with different numbers of
VGFs for different languages} (provided that there is at least one VGF
per language); in fact, the meta-level representations are simply
computed as the average of the views that are available for that
particular language. For reasons that will become clear in
Section~\ref{sec:ZSCLC}, this property allows \gfun\ to natively
operate in zero-shot mode.

In Section~\ref{sec:experiments:aggpolicies} we briefly report on some
preliminary experiments that we had carried out in order to assess the
relative merits of the two aggregation policies in terms of
classification performance. As we will see in
Section~\ref{sec:experiments:aggpolicies} in more detail, the results
of those experiments indicate that, while differences in performance
are small, they tend to be in favour of the averaging policy. This
fact, along with the fact that the averaging policy scales better with
the number of VGFs, and along with the fact that it allows different
numbers of VGFs for different languages, will eventually lead us to
opt for averaging as our aggregation policy of choice.

\subsection{Normalisation}
\label{sec:normalisation}

%
%


\noindent We have found that applying some routine normalisation
techniques to the vectors returned by our VGFs leads to consistent
performance improvements. This normalisation consists of the following
steps:

\begin{enumerate}

\item Apply (only for the MUSEs VGF and WCEs VGF) \emph{smooth inverse
  frequency} (SIF)~\cite{Arora:2017qq} to remove the first principal
  component of the document embeddings obtained as the weighted
  average of word embeddings. In their work,~\citet{Arora:2017qq} show
  that removing the first principal component from a matrix of
  document embeddings defined as a weighted average of word
  embeddings, is generally beneficial. The reason is that the way in
  which most word embeddings are trained tends to favour the
  accumulation of large components along semantically meaningless
  directions. However, note that for the MUSEs VGF and WCEs VGF we use
  the TFIDF weighting criterion instead of the criterion proposed
  by~\citet{Arora:2017qq}, since in our case we are modelling
  (potentially large) documents, instead of sentences like in their
  case.\footnote{The weighting technique proposed
  by~\citet{Arora:2017qq} does not account for term repetitions, since
  they make the assumption that words rarely occur more than once in a
  sentence. Conversely, when modelling entire documents, the TF factor
  may indeed play a fundamental role, and in such cases,
  as~\citet{Arora:2017qq} acknowledge, using TFIDF may be preferable.}

\item Impose unit L2-norm to the vectors before aggregating them by
  means of concatenation or averaging.

\item Standardize\footnote{Standardising (a.k.a.\ ``z-scoring'', or
  ``z-transforming'') consists of having a random variable $x$, with
  mean $\mu$ and standard deviation $\sigma$, translated and scaled as
  $z=\frac{x-\mu}{\sigma}$, so that the new random variable $z$ has
  zero mean and unit variance. The statistics $\mu$ and $\sigma$ are
  unknown, and are thus estimated on the training set.} the columns of
  the language-independent representations before training the
  classifiers (this includes (a) the classifiers in charge of
  homogenising the vector spaces before applying the averaging policy,
  and (b) the meta-classifier).

\end{enumerate}
\noindent The rationale behind these normalisation steps, when dealing
with heterogeneous representations, is straightforward and
two-fold. On one side, it is a means for equating the contributions
brought to the model by the different sources of information. On the
other, it is a way to counter the internal covariate shift across the
different sources of information (similar intuitions are well-known
and routinely applied when training deep neural architectures -- see,
e.g., \cite{Ioffe:2015sv}).

What might come as a surprise is the fact that normalisation helps
improve \gfun\ even when we equip \gfun\ only with the Posteriors VGF
(which coincides with the original \fun\ architecture), and that this
improvement is statistically significant. We quantify this variation
in performance in the experiments of Section~\ref{sec:experiments}.


\section{Experiments}
\label{sec:experiments}

\noindent In order to maximize the comparability with previous results
we adopt an experimental setting identical to the one used
in~\cite{Esuli:2019dp}, which we briefly sketch in this section. We
refer the reader to~\cite{Esuli:2019dp} for a more detailed discussion
of this experimental setting.


\subsection{Datasets}\label{sec:datasets}

\noindent The first of our two datasets is a version (created
by~\citet{Esuli:2019dp}) of RCV1/RCV2, a corpus of news stories
published by Reuters. This version of RCV1/RCV2 contains documents
each written in one of 9 languages (English, Italian, Spanish, French,
German, Swedish, Danish, Portuguese, and Dutch) and classified
according to a set of 73 classes. The dataset consists of 10 random
samples, obtained from the original RCV1/RCV2 corpus, each consisting
of 1,000 training documents and 1,000 test documents for each of the 9
languages (Dutch being an exception, since only 1,794 Dutch documents
are available; in this case, each sample consists of 1,000 training
documents and 794 test documents). \blue{\label{line:imbalance}Note
though that, while each random sample is balanced at the language
level (same number of training documents per language and same number
of test documents per language), it is not balanced at the class
level: at this level the dataset RCV1/RCV2 is highly imbalanced (the
number of documents per class ranges from 1 to 3,913 -- see
Table~\ref{tab:datasets}), and each of the 10 random samples is too.}
The fact that each language is equally represented in terms of both
training and test data allows the many-shot experiments to be carried
out in controlled experimental conditions, i.e., minimizes the
possibility that the effects observed for the different languages are
the result of different amounts of training data. (Of course,
zero-shot experiments will instead be run by excluding the relevant
training set(s).) Both the original RCV1/RCV2 corpus and the version
we use here are comparable at topic level, as news stories are not
direct translations of each other but simply discuss the same or
related events in different languages.

The second of our two datasets is a version (created
by~\citet{Esuli:2019dp}) of JRC-Acquis, a corpus of legislative texts
published by the European Union. This version of JRC-Acquis contains
documents each written in one of 11 languages (the same 9 languages of
RCV1/RCV2 plus Finnish and Hungarian) and classified according to a
set of 300 classes. The dataset is parallel, i.e., each document is
included in 11 translation-equivalent versions, one per language.
Similarly to the case of RCV1/RCV2 above, the dataset consists of 10
random samples, obtained from the original JRC-Acquis corpus,
\blue{each consisting of at least 1,000 training documents for each of
the 11 languages (summing up to a total of 12,687 training documents
in each sample), and 4,242 test documents for each of the 11
languages.} \blue{\label{line:imbalance2}As in the case of RCV1/RCV2,
this version of JRC-Acquis is not balanced at the class level (the
number of positive examples per class ranges from 55 to 1,155), and
the samples obtained from it are not balanced either.} Note that, in
this case, \citet{Esuli:2019dp} included at most one of the 11
language-specific versions in a training set, in order to avoid the
presence of translation-equivalent content in the training set; this
enables one to measure the contribution of training information coming
from different languages in a more realistic setting. When a document
is included in a test set, instead, all its 11 language-specific
versions are also included, in order to allow a perfectly fair
evaluation across languages, since each of the 11 languages is thus
evaluated on exactly the same content.

For both datasets, the results reported in this paper (similarly to
those of~\cite{Esuli:2019dp}) are averages across the 10 random
selections.
Summary characteristics of our two datasets are reported in
Table~\ref{tab:datasets}\blue{; \label{line:examples}excerpts from
sample documents from the two datasets are displayed in
Table~\ref{tab:examples}}.
%

\begin{table}[t]
  \begin{center}
    \centering \resizebox{\textwidth}{!} {
    \begin{tabular}{c|r|r|r|r|c|c|c|c|c|c}
      & \multicolumn{1}{c|}{$|\mathcal{L}|$}
      & \multicolumn{1}{c|}{$|\mathcal{Y}|$}
      & \multicolumn{1}{c|}{$|\mathrm{Tr}|$}
      & \multicolumn{1}{c|}{$|\mathrm{Te}|$}
      & \multicolumn{1}{c|}{Ave.Cls}
      & \multicolumn{1}{c|}{Min.Cls}
      & \multicolumn{1}{c|}{Max.Cls}
      & \multicolumn{1}{c|}{Min.Pos}
      & \multicolumn{1}{c|}{Max.Pos}
      & \multicolumn{1}{c}{Ave.Feats}
      \\
      \hline
      RCV1/RCV2 & 9 & 73 & 9,000 & \phantom{0}8,794 & 3.21 & 1 & 13 & \phantom{0}1 & 3,913 & 4,176 \\
      JRC-Acquis & 11 & 300 & 12,687 & 46,662 & 3.31 & 1 & 18 & 55 & 1,155 & 9,909 \\
    \end{tabular}
    }%
  \end{center}
  \caption{Characteristics of the datasets used in~\cite{Esuli:2019dp}
  and in this paper, including the number of languages
  ($|\mathcal{L}|$); number of classes ($|\mathcal{Y}|$); number of
  training ($|\mathrm{Tr}|$) and test ($|\mathrm{Te}|$) documents;
  average (Ave.Cls), minimum (Min.Cls), and maximum (Max.Cls) number
  of classes per document; minimum (Min.Pos) and maximum (Max.Pos)
  number of positive examples per class; and average number of
  distinct features per language (Ave.Feats).}
  \label{tab:datasets}
\end{table}

\begin{table}[t]
  \begin{center}
    \begin{tabular}{|p{0.7\linewidth} | p{0.3\linewidth}|}
      \hline
      \multicolumn{1}{|c|}{Text} & \multicolumn{1}{c|}{Labels} \\
      \hline\hline
      \footnotesize{\blue{BRAZIL: Talks stall on bill to
      scrap Brazil export tax. Voting to speed up a bill to remove a tax
      on Brazilian exports will take place August 27 at the earliest after
      federal and state governments failed to reach an accord on terms, a
      Planning Ministry spokeswoman said. Planning Minister Antonio Kandir
      and the Parana and Rio Grande do Sul governments have yet to agree
      on compensation following the proposed elimination of the so-called
      ICMS tax, which applies to products such as coffee, sugar and
      soyproducts. The elimination of the tax should inject at least
      \$1.5 billion into the agribusiness sector (...) \hfill [Other 505 words truncated]}}
                                 &
                                   \footnotesize{\blue{\begin{itemize} \item merchandise trade
                                     (E512)\item economics (ECAT)\item government finance (E21)\item trade/reserves
                                     (E51)\item expediture/revenue (E211) \end{itemize}
                                   \noindent 
                                   }}
      \\ 
      \hline
      \footnotesize{\blue{Commission Regulation (EC) No 1908/2004 of 29 October 2004 fixing
      the maximum aid for cream, butter and concentrated butter for the
      151th individual invitation to tender under the standing invitation
      to tender provided for in Regulation (EC) No 2571/97 THE COMMISSION
      OF THE EUROPEAN COMMUNITIES, Having regard to the Treaty
      establishing the European Community, Having regard to Council
      Regulation (EC) No 1255/1999 of 17 May 1999 on the common
      organisation of the market in milk and milk products [1], and in
      particular Article 10 thereof, Whereas: (1) The intervention
      agencies are, pursuant to Commission Regulation (EC) No 2571/97 of
      15 December 1997 on the sale of butter (...) \hfill [Other 243 words truncated]}}
                                 &
                                   \footnotesize{\blue{\begin{itemize} \item award of contract
                                     (20) \item concentrated product (2741) \item aid system (3003) \item farm price
                                     support (4236) \item butter (4860) \item youth movement (2004) \end{itemize}
                                   \noindent 
                                   }} \\
      \hline
    \end{tabular}
  \end{center}
    \caption{\blue{Excerpts from example documents from RCV1/RCV2 (1st
  example) and JRC-Acquis (2nd example).}}

  \label{tab:examples}
\end{table}


\subsection{Evaluation measures}

\noindent To assess the model performance we employ $F_{1}$, the
standard measure of text classification, and the more recently
theorized $K$ \cite{Sebastiani:2015zl}. These two functions are
defined as:
\begin{align}
  \label{eq:F1}
  F_{1} = & \ \left\{
            \begin{array}{cl}
              \dfrac{2\TP}{2\TP + \FP + \FN} & \mathrm{if} \ \TP + \FP + \FN>0 \rule[-3ex]{0mm}{7ex} \\
              1 & \mathrm{if} \ \TP=\FP=\FN=0 \\
            \end{array}
  \right.
\end{align}
\begin{align}
  \label{eq:K}
  K = & \ \left\{
        \begin{array}{cl}
          \dfrac{\TP}{\TP+\FN}+\dfrac{{\TN}}{{\TN}+\FP}-1 & \mathrm{if} \ \TP+\FN>0 \ \mathrm{and} \ \TN+\FP>0 \rule[-3ex]{0mm}{7ex} \\
          2\dfrac{{\TN}}{{\TN}+\FP}-1 & \mathrm{if} \ \TP+\FN=0 \rule[-3ex]{0mm}{7ex} \\
          2\dfrac{\TP}{\TP+\FN}-1 & \mathrm{if} \ \TN+\FP=0
        \end{array}
                                    \right.
\end{align}
\noindent where $\TP, \FP, \FN, \TN$ represent the number of true
positives, false positives, false negatives, and true negatives
generated by a binary classifier. $F_{1}$ ranges between 0 (worst) and
1 (best) and is the harmonic mean of precision and recall, while $K$
ranges between -1 (worst) and 1 (best).

To turn $F_{1}$ and $K$ (whose definitions above are suitable for
binary classification) into measures for multilabel classification, we
compute their microaverages ($F_{1}^{\mu}$ and $K^{\mu}$) and their
macroaverages ($F_{1}^{M}$ and $K^{M}$). $F_{1}^{\mu}$ and $K^{\mu}$
are obtained by first computing the class-specific values $\TP_{j}$,
$\FP_{j}$, $\FN_{j}$, $\TN_{j}$, computing
$\TP=\sum_{j=1}^{|\mathcal{Y}|}\TP_{j}$ (and analogously for
$\FP, \FN, \TN$), and then applying Equations~\ref{eq:F1} and
\ref{eq:K}. Instead, $F_{1}^{M}$ and $K^{M}$ are obtained by first
computing the class-specific values of $F_{1}$ and $K$ and then
averaging them across all $y_{j}\in\mathcal{Y}$.

We also test the statistical significance of differences in
performance 
via paired sample, two-tailed t-tests at the $\alpha=0.05$ and
$\alpha=0.001$ confidence levels.



\subsection{Learners}
\label{sec:learners}

\noindent Wherever possible, we use the same learner as used
in~\cite{Esuli:2019dp}, i.e., Support Vector Machines (SVMs) as
implemented in the \texttt{scikit-learn}
package.\footnote{\url{https://scikit-learn.org/stable/index.html}}
For the 2nd-tier classifier of \gfun, and for all the baseline
methods, we optimize the $C$ parameter, that trades off between
training error and margin, by testing all values $C=10^{i}$ for
$i\in\{-1, ..., 4\}$ by means of \label{line:5kfv}\blue{5-fold
cross-validation}.
We use Platt calibration in order to calibrate the 1st-tier
classifiers used in the Posteriors VGF and (when using averaging as
the aggregation policy) the classifiers that map document views into
vectors of posterior probabilities. We employ the linear kernel for
the 1st-tier classifiers used in the Posteriors VGF, and the RBF
kernel (i) for the classifiers used for implementing the averaging
aggregation policy, and (ii) for the 2nd-tier classifier.



In order to generate the BERT VGF (see Section~\ref{sec:vg_mbert}), we
rely on the pre-trained model released by
\texttt{Huggingface}\footnote{We use the
\texttt{bert-base-multilingual-cased} model available at
\url{https://huggingface.co/}} ~\cite{Wolf:2020ge}. For each run, we
train the model following the settings suggested by
\citet{Devlin:2019xx}, i.e., we add one classification layer on top of
the output of mBERT (the special token \texttt{[CLS]}) and fine-tune
the entire model end-to-end by minimising the binary cross-entropy
loss function. We use the AdamW optimizer~\cite{Loshchilov:2019dk}
with the learning rate set to 1e-5 and the weight decay set to
0.01. We also set the learning rate to decrease by means of a
scheduler (StepLR) with step size equal to 25 and gamma equal to
0.1. We set the training batch size to 4 and the maximum length of the
input (in terms of tokens) to 512 (which is the maximum input length
of the model). Given that the number of training examples in our
datasets is comparatively smaller than that used in
\citet{Devlin:2019xx}, we reduce the maximum number of epochs to 50,
and apply an early-stopping criterion that terminates the training
after 5 epochs showing no improvement (in terms of $F_{1}^{M}$) in the
validation set (a held-out split containing 20\% of the training
documents) in order to avoid overfitting. After convergence, we
perform one last training epoch on the validation set.

Each of the experiments we describe is performed 10 times, on 10
different samples extracted from the dataset, in order to assess its
statistical significance by means of the paired t-test mentioned in
Section~\ref{sec:normalisation}. All the results displayed in the
tables included in this paper are averages across these 10 samples and
across the $|\mathcal{L}|$ languages in the datasets.

We run all the experiments on a machine equipped with a 12-core
processor Intel Core i7-4930K at 3.40GHz with 32GB of RAM under Ubuntu
18.04 (LTS) and Nvidia GeForce GTX 1080 equipped with 8GB of RAM.


\subsection{Baselines}
\label{sec:baselines}

\noindent As the baselines against which to compare \gfun\ we use the
naïve monolingual baseline (hereafter indicated as \textsc{Na\"ive}),
Funnelling (\fun), plus the four best baselines
of~\cite{Esuli:2019dp}, namely, \emph{Lightweight Random Indexing}
(LRI~\cite{Moreo:2016fk}), \emph{Cross-Lingual Explicit Semantic
Analysis} (CLESA~\cite{Sorg:2012dn}), \emph{Kernel Canonical
Correlation Analysis} (KCCA~\cite{Vinokourov:2002mz}), and
\emph{Distributional Correspondence Indexing}
(DCI~\cite{Moreo:2016fg}). For all systems but \gfun, the results we
report are excerpted from~\cite{Esuli:2019dp}, so we refer to that
paper for the detailed setups of these baselines; the comparison is
fair anyway, since our experimental setup is identical to that
of~\cite{Esuli:2019dp}.

We also include 
mBERT~\cite{Devlin:2019xx} as an additional baseline. In order to
generate the mBERT baseline, we follow \blue{exactly the same}
procedure as described above for the BERT VGF.
\label{line:bert_difference} \blue{Note that the difference
between mBERT and BERT VGF comes down to the fact that the former
leverages a linear transformation of the document embeddings followed
by a sigmoid activation in order to compute the prediction scores. On
the other hand, BERT as a VGF is used as a feature extractor (or
embedder). Once the document representations are computed (by mBERT),
we project them to the space of the posterior probabilities via a set
of SVMs.}
We also experiment with an alternative training strategy in which we
simply train the classification layer, and leave the pre-trained
parameters of mBERT untouched, but omit the results obtained using
this strategy because in preliminary experiments it proved inferior to
the other strategy by a large margin.

Similarly to~\cite{Esuli:2019dp} we also report an ``idealized''
baseline (i.e., one whose performance all CLC methods should strive to
reach up to), called \textsc{UpperBound}, which consists of replacing
each non-English training example by its corresponding English
version, training a monolingual English classifier, and classifying
all the English test documents. \textsc{UpperBound} is present only in
the JRC-Acquis experiments since in RCV1/RCV2 the English versions of
non-English training examples are not available.


\subsection{Results of many-shot CLTC experiments}
\label{sec:cltc_results}

\noindent In this section we report the results that we have obtained
in our many-shot CLTC experiments on the RCV1/RCV2 and JRC-Acquis
datasets.\footnote{In an earlier, shorter version of this paper
\cite{Moreo:2021vn} we report different results for the very same
datasets. The reason of the difference is that in \cite{Moreo:2021vn}
we use concatenation as the aggregation policy while we here use
averaging.} These experiments are run in ``everybody-helps-everybody''
mode, i.e., all training data, from all languages, contribute to the
classification of all unlabelled data, from all languages.

We will use the notation \vgx\ to denote a \gfun\ instantiation that
uses only one VGF, namely the Posteriors VGF; \gfun\vgx\ is thus
equivalent to the original \fun\ architecture, but with the addition
of the normalisation steps discussed in
Section~\ref{sec:normalisation}. Analogously, \vgm\ will denote the
use of the MUSEs VGF (Section~\ref{sec:vg_muse}), \vgw\ the use of the
WCEs VGF (Section~\ref{sec:vg_wce}), and \vgb\ the use of the BERT VGF
(Section~\ref{sec:vg_mbert}).

Tables~\ref{tab:res_rcv} and~\ref{tab:res_jrc} report the results
obtained on RCV1/RCV2 and JRC-Acquis, respectively. We denote
different setups of \gfun\ by indicating after the hyphen the VGFs
that the variant uses. For each dataset we report the results for 7
different baselines and 9 different configurations of \gfun, as well
as for two distinct evaluation metrics ($F_{1}$ and $K$) aggregated
across the $|\mathcal{Y}|$ different classes by both micro- and
macro-averaging.

The results are grouped in four batches of methods. The first one
contains all baseline methods. The remaining batches present results
obtained using a selection of meaningful combinations of VGFs: the 2nd
batch reports the results obtained by \gfun\ when equipped with one
single VGF, the 3rd batch reports ablation results, i.e., results
obtained by removing one VGF 
from a setting containing all VGFs, while in
the last batch we report the results obtained by jointly using all the
VGFs discussed.

The results clearly indicate that the fine-tuned version of
multilingual BERT consistently outperforms all the other baselines, on
both datasets.
%
%
Concerning \gfun's results,
among the different settings of the second batch (testing different
VGFs in isolation), the only configuration that consistently
outperforms mBERT in RCV1/RCV2 is \gfun\vgb.
Conversely, on JRC-Acquis, all four VGFs in isolation manage to beat
mBERT for at least 2 evaluation measures. Most other configurations of
\gfun\ we have tested (i.e., configurations involving more than one
VGF) consistently beat mBERT, with the sole exception of \gfun\vgxmw\
on RCV1/RCV2.

\begin{table}[ht]
  \centering
  \begin{tabular}{lcccc}
    \toprule
    Method & $F_{1}^{M}$ & $F_{1}^{\mu}$ & $K^{M}$ & $K^{\mu}$\\
    \midrule
    \textsc{Na\"ive} & .467 $\pm$ .083\psdag & .776 $\pm$ .052\psdag & .417 $\pm$ .090\psdag & .690 $\pm$ .074\psdag \\

    \textsc{LRI}~\cite{Moreo:2016fk} & .490 $\pm$ .077\psdag & .771 $\pm$ .050\psdag & .440 $\pm$ .086\psdag & .696 $\pm$ .069\psdag \\
    \textsc{CLESA}~\cite{Sorg:2012dn} & .471 $\pm$ .074\psdag & .714 $\pm$ .061\psdag & .434 $\pm$ .080\psdag & .659 $\pm$ .075\psdag \\
    \textsc{KCCA}~\cite{Vinokourov:2002mz} & .385 $\pm$ .079\psdag & .616 $\pm$ .065\psdag & .358 $\pm$ .088\psdag & .550 $\pm$ .073\psdag \\
    \textsc{DCI}~\cite{Moreo:2016fg} & .485 $\pm$ .070\psdag & .770 $\pm$ .052\psdag & .456 $\pm$ .082\psdag & .696 $\pm$ .065\psdag \\

    FUN~\cite{Esuli:2019dp} & .534 $\pm$ .066\psdag & .802 $\pm$ .041\psdag & .506 $\pm$. 073\psdag & .760 $\pm$ .052\psdag \\
    mBERT~\cite{Devlin:2018dl} & \grayed{}.581 $\pm$ .014\psdag & \grayed{}.817 $\pm$ .005\psdag & \grayed{}.559 $\pm$ .015\psdag & \grayed{}.788 $\pm$ .008\psdag\\
    \hline
    \gfun-\vgx & .547 $\pm$ .065\psdag & .798 $\pm$ .041\psdag & .551 $\pm$ .070\psdag & \grayed{}.799 $\pm$ .046\psdag \\
    \gfun-\vgm & .548 $\pm$ .066\psdag & .769 $\pm$ .042\psdag & .564 $\pm$ .077\psdag & .765 $\pm$ .048\psdag \\
    \gfun-\vgw & .487 $\pm$ .062\psdag & .743 $\pm$ .054\psdag & .511 $\pm$ .086\psdag & .730 $\pm$ .058\psdag \\
    \gfun-\vgb				& \grayed{}.608 $\pm$ .064$^{\ddag}$ & \grayed{}.826 $\pm$ .040$^{\dag}$ & \grayed{}\textbf{.603} $\pm$ .078\psdag & .797 $\pm$ .049\psdag \\\hline

    \gfun-\vgxmb & \grayed{}\textbf{.611} $\pm$ .068\psdag & \grayed{}\textbf{.833} $\pm$ .035\psdag & \grayed{}.597 $\pm$ .077\sddag & \grayed{}\textbf{.813} $\pm$ .045\psdag \\
    \gfun-\vgxwb & .581 $\pm$ .062\psdag & .821 $\pm$ .037\psdag & .574 $\pm$ .073\psdag & .797 $\pm$ .046\psdag \\
    \gfun-\vgxmw & .558 $\pm$ .061\psdag & .801 $\pm$ .038\psdag & .558 $\pm$ .072\psdag & .788 $\pm$ .046\psdag \\
    \gfun-\vgwmb & .593 $\pm$ .065$^{\dag}$ & .821 $\pm$ .036\psdag & .582 $\pm$ .079\sdpsdag & .795 $\pm$ .048\psdag \\
    \hline

    \gfun-\vgxwmb & \grayed{}.596 $\pm$ .064$^{\dag}$ & \grayed{}.826 $\pm$ .035$^{\dag}$ & \grayed{}.579 $\pm$ .075\sdpsdag & \grayed{}.802 $\pm$ .046\psdag \\
    \bottomrule
  \end{tabular}
  \caption{Many-shot CLTC results on the RCV1/RCV2 dataset. Each cell
  reports the mean value and the standard deviation across the 10
  runs. \textbf{Boldface} indicates the best method overall, while
  greyed-out cells indicate the best method within the same group of
  methods. Superscripts $\dag$ and $\ddag$ denote the method (if any)
  whose score is not statistically significantly different from the
  best one; symbol $\dag$ indicates $0.001< p$-value $<0.05$ while
  symbol $\ddag$ indicates a $0.05\leq p$-value.}
  \label{tab:res_rcv}
\end{table}

\begin{table}[ht]
  \centering
  \begin{tabular}{lcccc}
    \toprule
    Method & $F_{1}^{M}$ & $F_{1}^{\mu}$ & $K^{M}$ & $K^{\mu}$\\
    \midrule
    \textsc{Na\"ive} & .340 $\pm$ .017\psdag & .559 $\pm$ .012\psdag & .288 $\pm$ .016\psdag & .429 $\pm$ .015\psdag \\
    \textsc{LRI}~\cite{Moreo:2016fk} & .411 $\pm$ .027\psdag & .594 $\pm$ .016\psdag & .348 $\pm$ .025\psdag & .476 $\pm$ .020\psdag \\
    \textsc{CLESA}~\cite{Sorg:2012dn} & .379 $\pm$ .034\psdag & .557 $\pm$ .024\psdag & .330 $\pm$ .034\psdag & .453 $\pm$ .029\psdag \\
    \textsc{KCCA}~\cite{Vinokourov:2002mz} & .206 $\pm$ .018\psdag & .357 $\pm$ .023\psdag & .176 $\pm$ .017\psdag & .244 $\pm$ .022\psdag \\
    \textsc{DCI}~\cite{Moreo:2016fg} & .317 $\pm$ .012\psdag & .510 $\pm$ .014\psdag & .274 $\pm$ .013\psdag & .382 $\pm$ .016\psdag \\

    \fun~\cite{Esuli:2019dp} & .399 $\pm$ .013\psdag & .587 $\pm$ .009\psdag & .365 $\pm$ .014\psdag & .490 $\pm$ .013\psdag \\
    mBERT~\cite{Devlin:2018dl} & \grayed{}.420 $\pm$ .023\psdag & \grayed{}.608 $\pm$ .016\psdag & \grayed{}.379 $\pm$ .006\psdag & \grayed{}.507 $\pm$ .009\psdag \\
    \hline
    \gfun-\vgx & .432 $\pm$ .015\psdag & .587 $\pm$ .010\psdag & .441 $\pm$ .016\psdag & .553 $\pm$ .013\psdag \\
    \gfun-\vgm & .440 $\pm$ .039\psdag & .586 $\pm$ .032\psdag & .442 $\pm$ .045\psdag & .549 $\pm$ .034\psdag \\
    \gfun-\vgw & .410 $\pm$ .016\psdag & .553 $\pm$ .014\psdag & .410 $\pm$ .021\psdag & .525 $\pm$ .022\psdag \\
    \gfun-\vgb & \grayed{}.501 $\pm$ .023\psdag & \grayed{}.627 $\pm$ .016\psdag & \grayed{}.485 $\pm$ .023\psdag & \grayed{}.574 $\pm$ .019\psdag \\\hline

    \gfun-\vgxmb 	& \grayed{}\textbf{.525} $\pm$ .020\psdag & \grayed{}\textbf{.649} $\pm$ .014\psdag & \grayed{}\textbf{.528} $\pm$ .023\psdag & \grayed{}\textbf{.620} $\pm$ .017\psdag \\
    \gfun-\vgxwb 	& .497 $\pm$ .011\psdag & .621 $\pm$ .008\psdag & .508 $\pm$ .011\psdag & .606 $\pm$ .010\psdag \\
    \gfun-\vgxmw 	& .475 $\pm$ .012\psdag & .604 $\pm$ .010\psdag & .489 $\pm$ .014\psdag & .593 $\pm$ .011\psdag \\
    \gfun-\vgwmb 	& .513 $\pm$ .016\psdag & .632 $\pm$ .011\psdag & .522 $\pm$ .017\sddag & .619 $\pm$ .013\sddag \\
    \hline

    \gfun-\vgxwmb & \grayed{}.514 $\pm$ .014\psdag & \grayed{}.635 $\pm$ .010\psdag & \grayed{}.521 $\pm$ .015\sdpsdag & \grayed{}.618 $\pm$ .011\sddag \\
    \hline
    \textsc{UpperBound} & \multicolumn{1}{l}{.599} & \multicolumn{1}{l}{.707} & \multicolumn{1}{l}{.547} & \multicolumn{1}{l}{.632} \\
    \bottomrule
  \end{tabular}
  \caption{As Table~\ref{tab:res_rcv}, but using JRC-Acquis instead of
  RCV1/RCV2.}
  \label{tab:res_jrc}
\end{table}



Something that jumps to the eye is that \gfun\vgx\ yields better
results than \fun, which is different from it only for the the
normalisation steps of Section~\ref{sec:normalisation}. This is a
clear indication that these normalisation steps are indeed beneficial.

Combinations relying on WCEs seem to perform comparably better in the
JRC-Acquis dataset, and worse in RCV1/RCV2. This can be ascribed to
the fact that the amount of information brought about by word-class
correlations is higher in the case of JRC-Acquis (since this dataset
contains no fewer than 300 classes) than in RCV1/RCV2 (which only
contains 73 classes). Notwithstanding this, the WCEs VGF seems to be
the weakest among the VGFs that we have tested. Conversely, the
strongest VGF seems to be the one based on mBERT, though it is also
clear from the results that other VGFs contribute to further improve
the performance of \gfun; in particular, the combination \gfun\vgxmb\
stands as the top performer overall, since it is always either the
best performing model or a model no different from the best performer
in a statistically significant sense.

\blue{\label{line:singlacontribution} Upon closer examination of
Tables \ref{tab:res_rcv} and \ref{tab:res_jrc}, the 2nd, 3rd, and 4th
batches help us in highlighting the contribution of each signal (i.e.,
information brought about by the VGFs).

Let us start from the 4th batch, where we report the results obtained
by the configuration of \gfun\ that exploits all of the available
signals (\gfun-XWMB). In RCV1/RCV2 such a configuration yields
superior results to the single-VGF settings (note that even though
results for \gfun-B (.608) are higher than those for \gfun-XWMB
(.596), this difference is not statistically significant, with a
$p$-value of .680, according to the two-tailed t-test that we have
run). Such a result indicates that there is indeed a synergy among the
heterogeneous representations.

In the 3rd batch, we investigate whether all of the signals are
mutually beneficial or if there is some redundancy among them. We
remove from the ``full stack'' (\gfun\vgxwmb) one VGF at a time. The
removal of the BERT VGF has the worst impact on $F_{1}^{M}$. This was
expected since, in the single-VGF experiments, \gfun-B was the
top-performing setup.  Analogously, by removing representations
generated by the Posteriors VGF or those generated by the MUSEs VGF,
we have a smaller decrease in $F_{1}^{M}$ results. On the contrary,
ditching WCEs results in a higher $F_{1}^{M}$ score (our top-scoring
configuration); the difference between \gfun\vgxwmb\ and \gfun\vgxmb\
is not statically significant in RCV1/RCV2 (with a $p$-value between
0.001 and 0.05), but it is significant in JRC-Acquis. This is an
interesting fact: despite the fact that in the single-VGF setting the
WCEs VGF is the worst-performing, we were not expecting its removal to
be beneficial. Such a behaviour suggests that the WCEs are not
well-aligned with the other representations, resulting in worse
performance across all the four metrics.  This is also evident if we
look at results reported in \cite{Pedrotti:2020pd}. If we remove from
\gfun\vgxmw\ (.558) the Posteriors VGF, thus obtaining \gfun\vgmw, we
obtain a $F_{1}^{M}$ score of .536; by removing the MUSEs VGF, thus
obtaining \gfun-XW, we lower the $F_{1}^{M}$ to .523; instead, by
discarding the WCEs VGF, thus obtaining \gfun-XM, we increase
$F_{1}^{M}$ to .575. This behaviour tells us that the information
encoded in the Posteriors and WCEs representations is diverging: in
other words, it does not help in building more easily separable
document embeddings. Results on JRC-Acquis are along the same line.}

\begin{figure}[h] \centering
  \includegraphics[width=1.\textwidth]{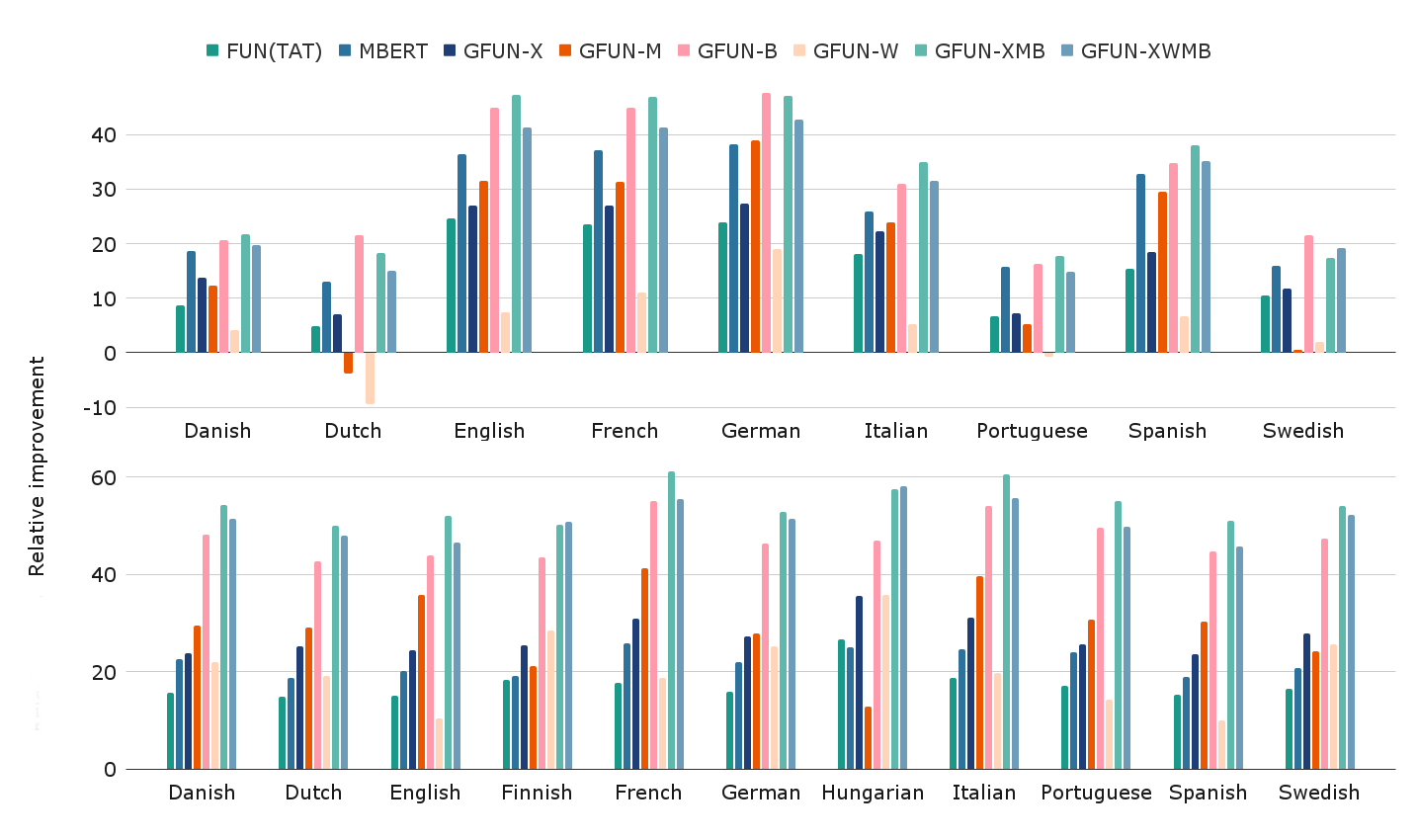}
 \caption{
 Percentage of relative improvement per language obtained by different
 cross-lingual models in the many-shot CLTC experiments, in terms of
 $F_{1}^{M}$ with respect to the \textsc{Na\"ive} solution, for
 RCV1/RCV2 (top) and JRC-Acquis (bottom).}
 \label{fig:lang_comparison_rcv}
\end{figure}

In Figure~\ref{fig:lang_comparison_rcv}, we show a more in-depth
analysis of the results,
in which we compare, for each language, the relative improvements
obtained in terms of $F_{1}^{M}$ (the other evaluation measures show
similar patterns) by mBERT (the top-performing baseline) and a
selection of \gfun\ configurations, with respect to the
\textsc{Na\"ive} solution.

These results confirm that the improvements brought about by
\gfun\vgx\ with respect to \fun\ are consistent across all languages,
and not only as an average across them, for both datasets.
The only configurations that underperform some monolingual na\"ive
solutions (i.e., that have a \textit{negative} relative improvement)
are \gfun\vgm\ (for Dutch) and \gfun\vgw\ (for Dutch and Portuguese)
on RCV1/RCV2. These are also the only configurations that sometimes
fare worse than the original \fun.
The configurations \gfun\vgb, \gfun\vgxmb, and \gfun\vgxwmb, all
perform better than the baseline mBERT on almost all languages and on
both datasets (the only exception for this happens for Portuguese when
using \gfun\vgxwmb\ in RCV1/RCV2), with the improvements with respect
to mBERT being markedly higher on JRC-Acquis.
Again, we note that, despite the clear evidence that the VGF based on
mBERT brings to bear the highest improvements overall, all other VGFs
do contribute to improving the classification performance; the
histograms of Figure~\ref{fig:lang_comparison_rcv} now reveal that the
contributions are consistent across all languages. For example,
\gfun\vgxmb\ outperforms \gfun\vgb\ for six out of nine languages in
RCV1/RCV2, and for all eleven languages in JRC-Acquis.

\blue{\label{line:independence}As a final remark, we should note
that the document representations generated by the different VGFs are
certainly not entirely independent (although their degree of mutual
dependence would be hard to measure precisely), since they are all
based on the distributional hypothesis, i.e., on the notion that
systematic co-occurrence (of words and other words, of words and
classes, of classes and other classes, etc.) is an evidence of
correlation. However, in data science, mutual independence is not a
necessary condition for usefulness; we all know this, e.g., from the
fact that the ``bag of words'' model of representing text works well
despite the fact that it makes use of thousands of features that are
not independent of each other.
Our results show that, in the best-performing setups of \gfun, several
such VGFs coexist despite the fact that they are probably not mutually
independent, which seems to indicate that the lack of independence of
these VGFs is not an obstacle.
}


\subsection{Results of zero-shot CLTC experiments}
\label{sec:ZSCLC}

\noindent \fun\ was not originally designed for dealing with zero-shot
scenarios since, in the absence of training documents for a given
language, the corresponding first-tier language-dependent classifier
cannot be trained. Nevertheless,~\citet{Esuli:2019dp} managed to
perform zero-shot cross-lingual experiments by plugging in an
auxiliary classifier trained on MUSEs representations
that is invoked for any target language for which training data are
not available, provided that this language is among the 30 languages
covered by MUSEs.

Instead, \gfun\ caters for zero-shot cross-lingual classification
\textit{natively}, provided that at least one among the VGFs it uses
is able to generate representations for the target language with no
training data (for the VGFs described in this paper, this is the case
of the MUSEs VGF and mBERT VGF for all the languages they cover). To
see why, assume the \gfun\vgxwmb\ instance of \gfun\ using the
averaging procedure for aggregation (Section \ref{sec:aggfun}). Assume
that there are training documents for English, and that there are no
training data for Danish. We train the system in the usual way
(Section \ref{sec:themethod}).
For a Danish test document, the MUSEs VGF\footnote{\label{foot:idf}In
the absence of a proper training set, the IDF factor needed for
computing the TFIDF weighting can be estimated using the test
documents themselves, since TFIDF is an unsupervised weighting
function.} and the mBERT VGF contribute to its representation, since
Danish is one of the languages covered by MUSEs and mBERT. The
aggregation function averages across all four VGFs (\vgxwmb) for
English test documents, while it only averages across two VGFs (-MB)
for Danish test documents. Note that the meta-classifier does not
perceive differences between English test documents and Danish test
documents since, in both cases, the representations it receives from
the first tier come down to averages of calibrated (and normalized)
posterior probabilities. Therefore, any language for which there are
no training examples can be dealt with by our instantiation of \gfun\
provided that this language is catered for by MUSEs and/or mBERT.

To obtain results directly comparable with the zero-shot setup
employed by~\citet{Esuli:2019dp}, we reproduce their experimental
setup. Thus, we run experiments in which we start with one single
source language (i.e., a language endowed with its own training data),
and we add new source languages iteratively, one at a time (in
alphabetical order), until all languages for the given dataset are
covered. At each iteration, we train \gfun\ on the available source
languages, and test on \textit{all} the target languages. At the
$i$-th iteration we thus have $i$ source languages and $|\mathcal{L}|$
target (test) languages, among which $i$ languages have their own
training examples and the other $(|\mathcal{L}|-i)$ languages do not.
For this experiment we choose the configuration involving all the VGFs
(\gfun\vgxwmb).

\begin{figure}[ht!] \centering
  \includegraphics[height=.91\textheight]{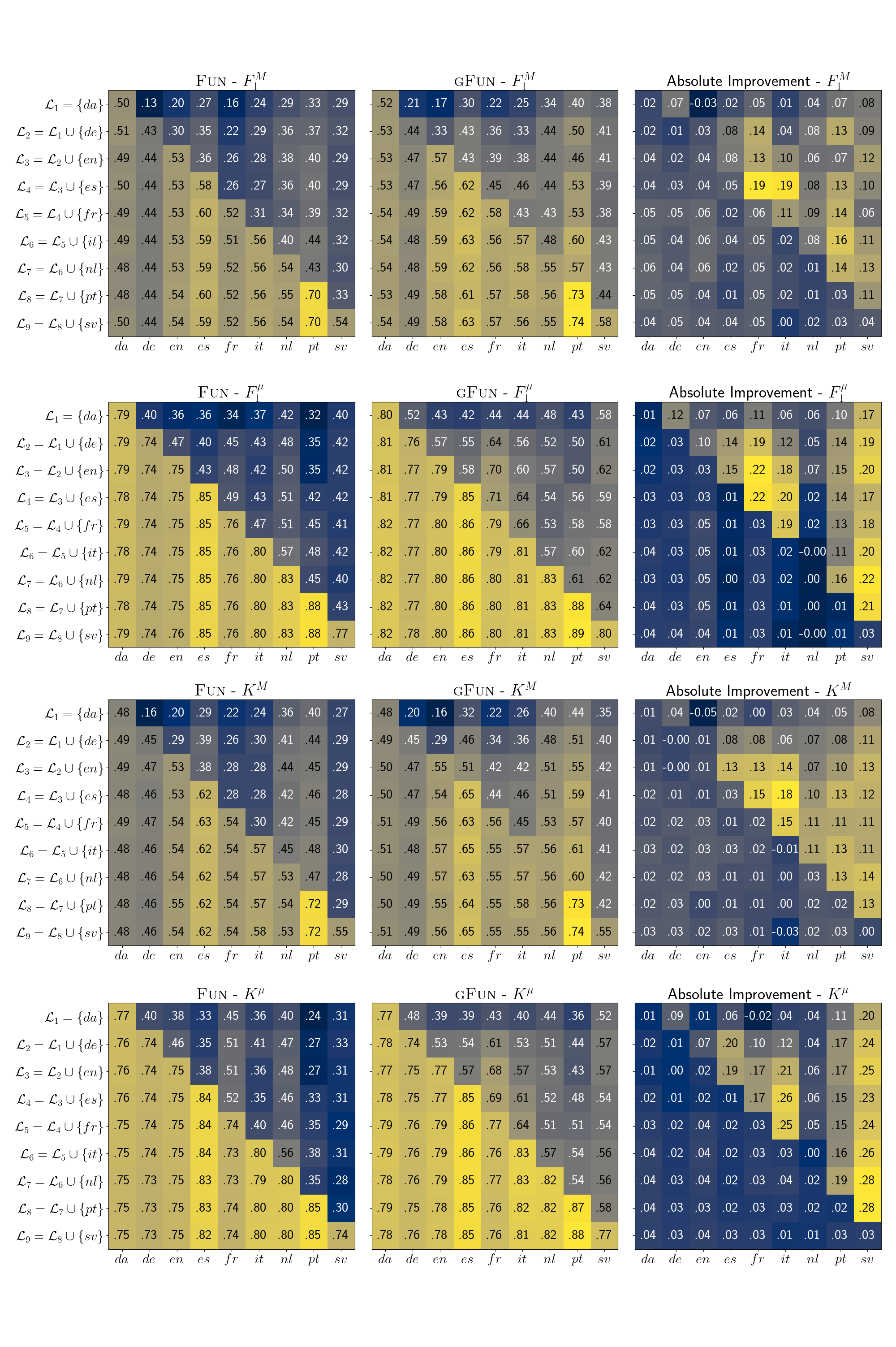}
 \caption{
 Results of zero-shot CLTC experiments on RCV1/RCV2 
 }
 \label{fig:zscl_rcv}
\end{figure}

\begin{figure}[ht!] \centering
  \includegraphics[height=.91\textheight]{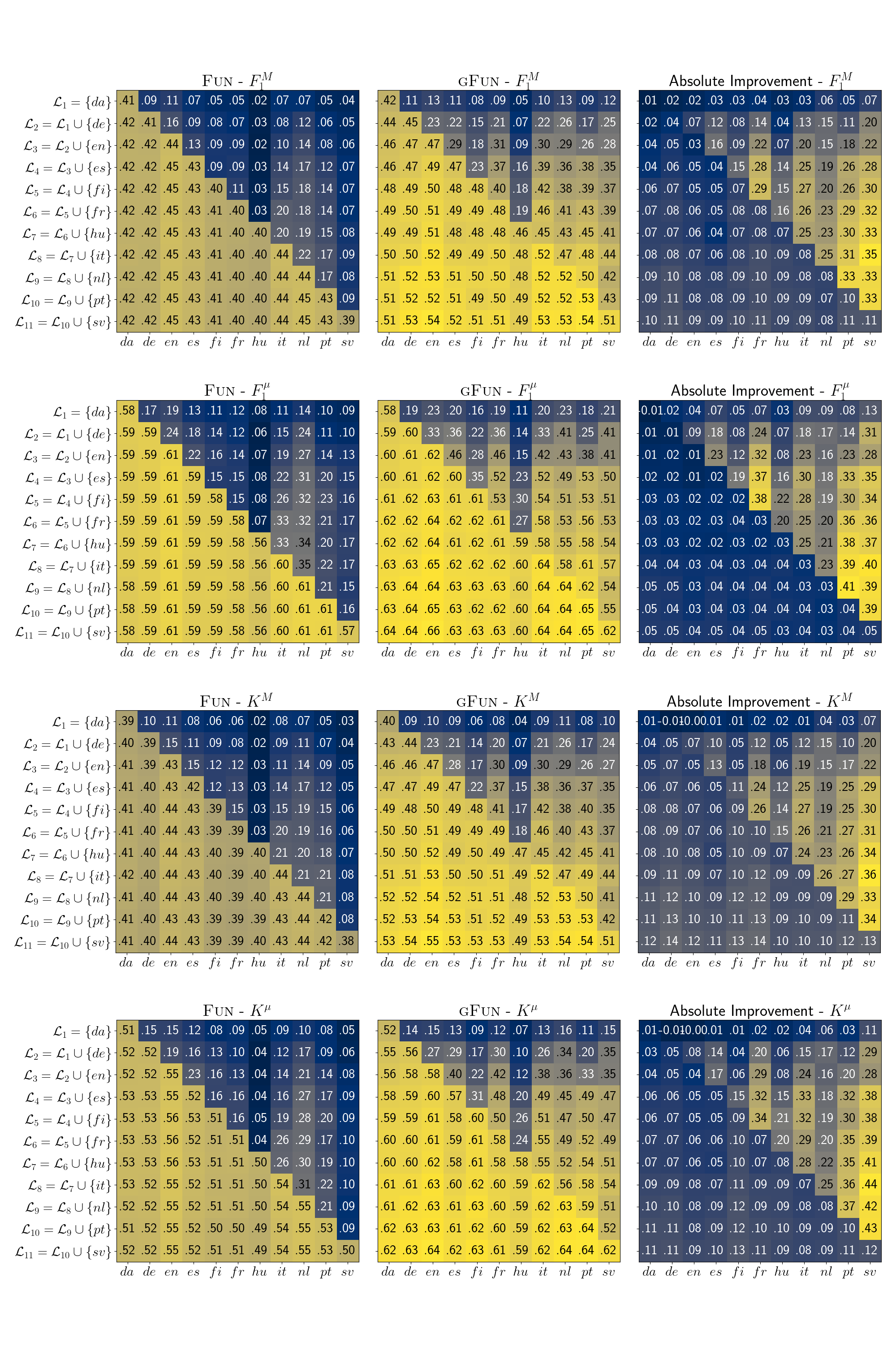}
 \caption{
 Results of zero-shot CLTC experiments on JRC-Acquis 
 }
 \label{fig:zscl_jrc}
\end{figure}

%
The results are reported in Figure \ref{fig:zscl_rcv} and
Figure~\ref{fig:zscl_jrc}, where we compare the results obtained by
\fun\ and \gfun\vgxwmb\ on both datasets, for all our evaluation
measures. Results are presented in a grid of three columns, in which
the first one corresponds to the results of \fun\ as reported
in~\cite{Esuli:2019dp}, the second one corresponds to the results
obtained by \gfun\vgxwmb, and the third one corresponds to the
difference between the two, in terms of absolute improvement of
\gfun\vgxwmb\ w.r.t. \fun. The results are arranged in four rows, one
for each evaluation measure. Performance scores are displayed through
heat-maps, in which columns represent target languages, and rows
represent training iterations (with incrementally added source
languages). Colour coding helps interpret and compare the results: we
use red for indicating low values of accuracy and green for indicating
high values of accuracy (according to the evaluation measure used) for
the first and second columns; the third column (absolute improvement)
uses a different colour map, ranging from dark blue (low improvement)
to light green (high improvement). The tone intensities of the \fun\
and \gfun\ colour maps for the different evaluation measures are
independent of each other, so that the darkest red (resp., the
lightest green) always indicates the worst (resp., the best) result
obtained by any of the two systems \textit{for the specific evaluation
measure}.

%

Note that the lower triangular matrix within each heat map reports
results for standard (many-shot) cross-lingual experiments, while all
entries above the main diagonal report results for zero-shot
cross-lingual experiments. As was to be expected, results for
many-shots experiments tend to display higher figures (i.e., greener
cells), while results for zero-shot experiments generally display
lower figures (i.e., redder cells).
These figures clearly show the superiority of \gfun\ over \fun, and
especially so for the zero-shot setting, for which the magnitude of
improvement is decidedly higher. The absolute improvement ranges from
18\% of $K^{M}$ to 28\% of $K^{\mu}$ on RCV1/RCV2, and from 35\% of
$F_{1}^{M}$ to 44\% of $K^{\mu}$ in the case of JRC-Acquis.

In both datasets, the addition of new languages to the training set
tends to help \gfun\ improve the classification of test documents also
for other languages for which a training set was already available
anyway. This is witnessed by the fact that the green tonality of the
columns in the lower triangular matrix becomes gradually darker; for
example, in JRC-Acquis, the classification of test documents in Danish
evolves stepwise from $K=0.52$ (when the training set consists only of
Danish documents) to $K=0.62$ (when all languages are present in the
training set).\footnote{That the addition of new languages to the
training set helps improve the classification of test documents for
other languages for which a training set was already available, is
true also in \fun. However, this does not emerge from Figure
\ref{fig:zscl_rcv} and Figure~\ref{fig:zscl_jrc} (which are taken from
\cite{Esuli:2019dp}). This has already been noticed by
\citet{Esuli:2019dp}, who argue that this happens only in the
zero-shot version of \fun, and is due to the
zero-shot classifier's failure to deliver well calibrated
probabilities.}


\begin{figure}[th] \centering
  \includegraphics[width=.95\textwidth]{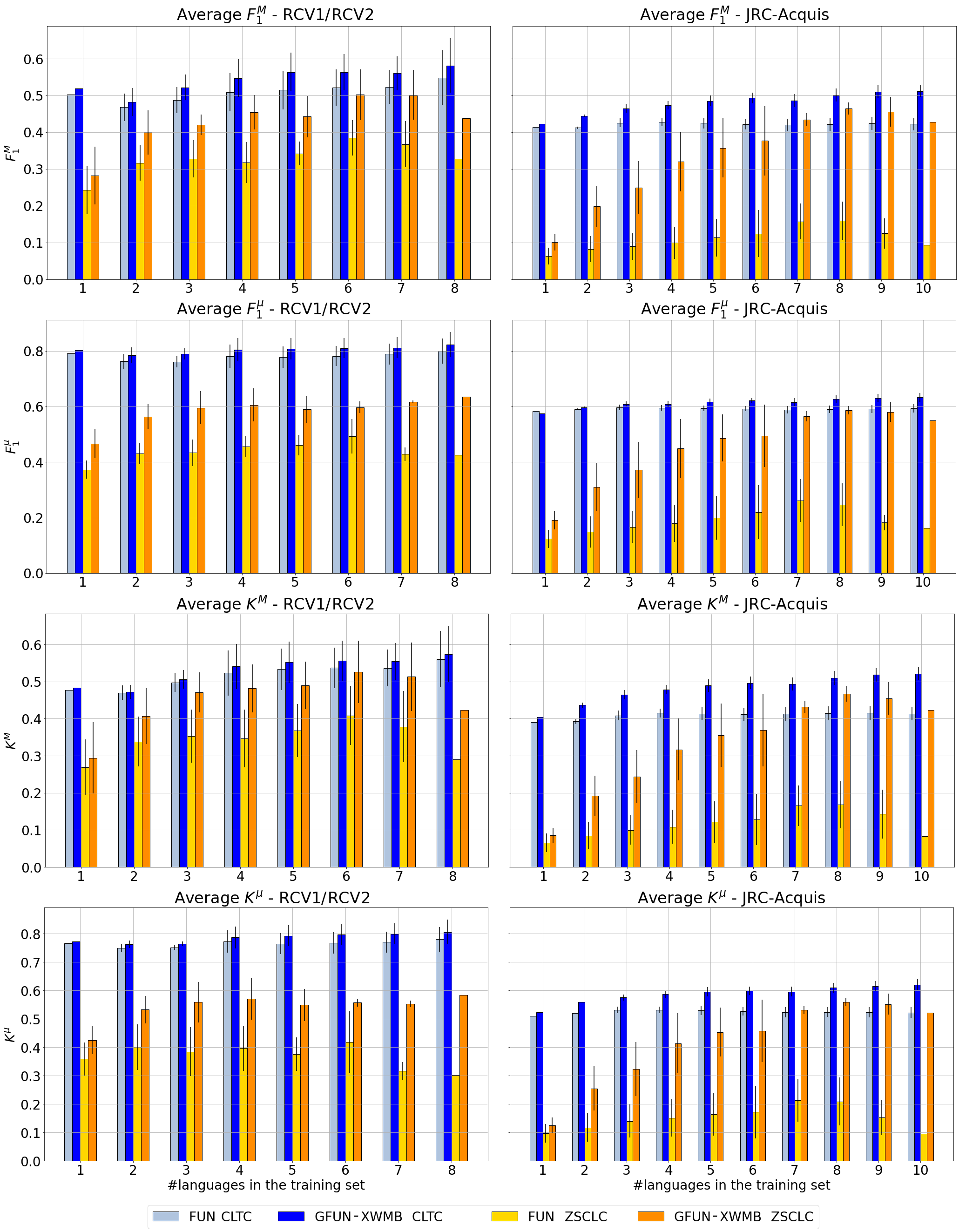}
  \caption{Performance of different CLTC systems as a function of the
  number of language-specific training sets used.}
  \label{fig:averagev_cltc}
\end{figure}

A direct comparison between the old and new variants of funnelling is
conveniently summarized in
Figure~\ref{fig:averagev_cltc}, where we display average values of
accuracy (in terms of our four evaluation measures) obtained by each
method across all experiments of the same type, i.e., standard
cross-lingual (CLTC -- values from the lower diagonal matrices of
Figures~\ref{fig:zscl_rcv} and~\ref{fig:zscl_jrc}) or zero-shot
cross-lingual (ZSCLC -- values from the upper diagonal matrices), as a
function of the number of training languages, for both datasets.
These histograms reveal that \gfun\ improves over \fun\
in the zero-shot experiments. Interestingly enough, the addition of
languages to the training set seems to have a positive impact in
\gfun, both for zero-shot and cross-lingual experiments.

\subsection{Testing different aggregation policies}
\label{sec:experiments:aggpolicies}

\noindent In this brief section we summarize the results of
preliminary, extensive experiments in which we had compared the
performance of different aggregation policies (concatenation vs.\
averaging); we here report only the results for the \gfun\vgxm\ and
\gfun\vgxmw\ models (the complete set of experiments is described
in~\cite{Pedrotti:2020pd}).

Table~\ref{tab:comparison_aggregation}
reports the results we obtained for RCV1/RCV2 and JRC-Acquis,
respectively. The results conclusively indicate that the averaging
aggregation policy yields either the best results, or results that are
not different (in a statistically significant sense) from the best
ones, in all cases. This, along with other motivations discussed in
Section~\ref{sec:aggfun} (scalability, and the fact that it enables
zero-shot classification) makes us lean towards adopting averaging as
the default aggregation policy.

\blue{\label{line:moreinfo}Incidentally,
Table~\ref{tab:comparison_aggregation} also seems to indicate that
WCEs work better in JRC-Acquis than in RCV1/RCV2. This is likely due
to the fact that, as observed in \cite{Moreo:2021qq}, the benefit
brought about by WCEs tends to be more substantial when the number of
classes is higher, since a higher number of classes means that WCEs
have a higher dimensionality, and that they thus bring more
information to the process.}

\begin{table}[t]
  \resizebox{\textwidth}{!}{
  \begin{tabular}{ll|cccc|cccc}
    \toprule
    & & \multicolumn{4}{c}{RCV1/RCV2} & \multicolumn{4}{|c}{JRC-Acquis} \\\hline
    Method & Policy & $F_{1}^{M}$ & $F_{1}^{\mu}$ & $K^{M}$ & $K^{\mu}$ & $F_{1}^{M}$ & $F_{1}^{\mu}$ & $K^{M}$ & $K^{\mu}$ \\
    \hline

    \gfun\vgxm & Concatenation & 0.562$^{\ddag}$ & \textbf{0.806}\psdag & 0.552$^\dag$ & 0.797$^{\ddag}$ 
                                  & 0.468\psdag & 0.610\psdag & 0.466\psdag & 0.572\psdag \\
 
    \gfun\vgxm & Averaging & \textbf{0.573}\psdag & 0.805$^{\ddag}$ & \textbf{0.575}\psdag & \textbf{0.800}\psdag & \textbf{0.477}\psdag & \textbf{0.615}\psdag & 0.488$^{\ddag}$ & 0.588\psdag \\
 
    \gfun\vgxmw & Concatenation & 0.540\psdag & 0.791\psdag & 0.530\psdag & 0.773\psdag & 0.461\psdag & 0.609\psdag & 0.445\psdag & 0.560\psdag \\
 
    \gfun\vgxmw & Averaging & 0.558$^\dag$ & 0.801$^\dag$ & 0.558$^\dag$ & 0.788\psdag & 0.475$^{\ddag}$ & 0.604\psdag & \textbf{0.489}\psdag & \textbf{0.593}\psdag \\
    \bottomrule
  \end{tabular}
  }%
  \caption{Results of many-shot CLTC experiments comparing the two
  aggregation policies on RCV1/RCV2 and JRC-Acquis
  (from~\cite{Pedrotti:2020pd}).}
  \label{tab:comparison_aggregation}
\end{table}


\subsection{Learning-Curve Experiments}\label{sec:further_experiments}


\blue{In this section we report the results obtained in additional
experiments aiming to quantify the impact on accuracy of variable
amounts of target-language training documents. Given the supplementary
nature of these experiments, we limit them to the RCV1/RCV2
dataset. Furthermore, for computational reasons we carry out these
experiments only on a subset of the original languages (namely,
English, German, French, and Italian).}
\blue{
In Figure \ref{fig:low_regime} we report the results, in terms of
$F^1_M$, obtained on RCV1/RCV2. For each of the 4 languages we work
on, we assess the performance of \gfun\vgxmb\ by varying the amount of
target-language training documents; we carry out experiments with 0\%,
10\%, 20\%, 30\%, 50\%, and 100\% of the training documents. For
example, the experiments on French (Figure \ref{fig:low_regime},
bottom left) are run by testing on 100\% of the French test data a
classifier trained with 100\% of the English, German, and Italian
training data and with variable proportions of the French training
data. We compare the results with those obtained (using the same
experimental setup) by the Na\"ive approach (see Section
\ref{sec:introduction} and \ref{sec:datasets}) and by \fun
\citep{Esuli:2019dp}.

\begin{figure}[ht] \centering
  \includegraphics[width=.95\textwidth]{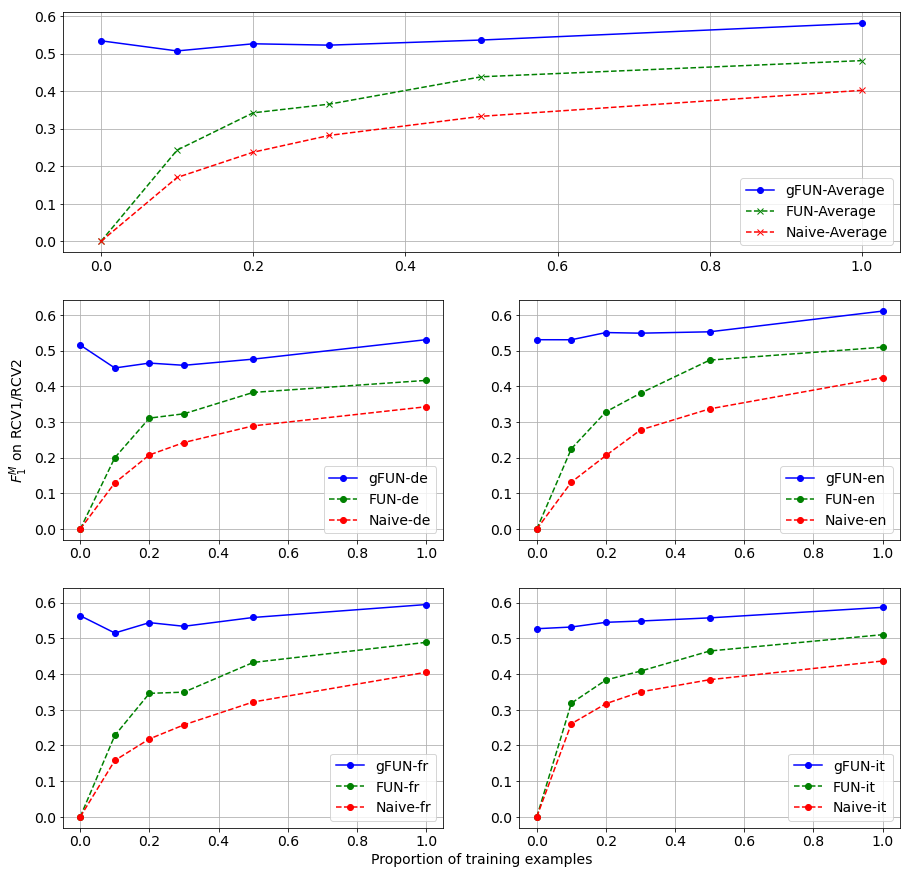}
  \caption{\blue{Learning-curve experiments performed on RCV1/RCV2
  dataset. Experiments are performed for increasing proportions of
  training examples (i.e., for $0\%, 10\%, 20\%, 30\%, 50\%, 100\%$)
  for four languages (i.e., German, English, French, and Italian).
  The configuration of \gfun\ deployed is \gfun\vgxmb. We compare the
  performance of \gfun\vgxmb\ with that displayed by FUN
  \citep{Esuli:2019dp} and by the Na\"ive approach.}}
  \label{fig:low_regime}
\end{figure}
It is immediate to note from the plots that the two baseline systems
have a very low performance when there are few target-language
training examples, but this is not true for \gfun\vgwmb, which has a
very respectable performance even with 0\% target-language training
examples; indeed, \gfun\vgwmb\ is able to almost bridge the gap
between the zero-shot and many-shot settings, i.e., for \gfun\vgwmb\
the difference between the $F^1_M$ values obtained with 0\% or 100\%
target-language training examples is moderate. On the contrary, for
the two baseline systems considered, the inclusion of additional
target-language training examples results in a substantial increase in
performance; however, both baselines substantially underperform
\gfun\vgwmb, for any percentage of target-language training examples,
and for each of the 4 target languages.
}





\section{Learning alternative composition functions: The Recurrent
VGF}
\label{sec:further}

%

\noindent 
The embeddings-based VGFs that we have described in
Sections~\ref{sec:vg_muse} and~\ref{sec:vg_wce} implement a simple dot
product as a means for deriving document embeddings from the word
embeddings and the TFIDF-weighted document vector. However, while such
an approach is known to produce document representations that perform
reasonably well on short texts~\cite{de2016representation}, there is
also evidence that more powerful models are needed for learning more
complex ``composition functions'' for
texts~\cite{Dasgupta:2018rt,socher2011parsing}. In NLP and related
disciplines, \emph{composition functions} are defined as functions
that take as input the constituents of a sentence (sometimes already
converted into distributed dense representations), and output a single
vectorial representation capturing the overall semantics of the given
sentence. In this section, we explore alternatives to the dot product
for the VGFs based on MUSEs and WCE.

For this experiment, for generating document embeddings we rely on
recurrent neural networks (RNNs). In particular, we adopt the
\textit{gated recurrent unit} (GRU)~\cite{Cho:2014zc}, a lightweight
variant of the \textit{long-short term memory} (LSTM)
unit~\cite{Hochreiter:1997fv}, as our recurrent cell. GRUs have fewer
parameters than LSTMs and do not learn a separate output function
(such as the output gate in LSTMs), and are thus more efficient during
training. (In preliminary experiments we have carried out, we have
found no significant differences in performance between GRU and LSTM;
the former is much faster to train, though.) This gives rise to what
we call the \textit{Recurrent VGF}.

In the Recurrent VGF we thus infer the composition function
at VGF fitting time. During the training phase, we train an RNN to
generate good document representations from a set of language-aligned
word representations consisting of the concatenation of WCEs and
MUSEs.
This VGF is trained in an end-to-end fashion. The output
representations of the training documents generated by the GRU are
projected onto a $|\mathcal{Y}|$-dimensional space of label
predictions; the network is trained by minimising the binary
cross-entropy loss between the predictions and the true labels. We
explore different variants depending on how the parameters of the
embedding layer are initialized (see below). We do not freeze the
parameters of the embedding layers, so as to allow the optimisation
procedure to fine-tune the embeddings.
We use the Adam optimizer~\cite{Kingma:2015ol} with initial learning
rate set at 1e-3 and no weight decay. We halve the learning rate every
25 epochs by means of StepLR (gamma = 0.5, step size = 25). We set the
training batch size to 256 and compute the maximum length of the
documents dynamically at each batch by taking their average
length. \blue{\label{line:doc_length}Documents exceeding the
computed length are truncated, whereas shorter ones are padded.}
Finally, we train the model for a maximum of 250 epochs, with an
early-stopping criterion that terminates the training after 25 epochs
with no improvement on the validation $F_{1}^{M}$.

There is only one Recurrent VGF in the entire \gfun\ architecture,
which processes all documents, independently of the language they
belong to.
Once trained, the last linear layer is discarded. All training
documents are then passed through the GRU and converted into document
embeddings, which are eventually used to train a calibrated classifier
which returns posterior probabilities for each class in the codeframe.




\subsection{Experiments}
\label{sec:recurrentvgfexperiments}

\noindent We perform many-shot CLTC experiments using the Recurrent
VGF trained on MUSEs only (denoted \vggM), or trained on the
concatenation of MUSEs and WCEs (denoted \vggMW). We do not explore
the case in which the GRU is trained exclusively on WCEs since, as
explained in~\cite{Moreo:2021qq}, WCEs are meant to be concatenated to
general-purpose word embeddings. Similarly, we avoid exploring
combinations of VGFs based on redundant sources of information, e.g.,
we do not attempt to combine the MUSEs VGFs with the Recurrent VGF,
since this latter already makes use of MUSEs. 

Tables \ref{tab:res_rcv:rec} and \ref{tab:res_jrc:rec} report on the
experiments we have carried out using the Recurrent VGF, in terms of
all our evaluation measures, for RCV1/RCV2 and JRC-Acquis,
respectively. These results indicate that the Recurrent VGF
under-performs the dot product criterion (this can be easily seen by
comparing each result with its counterpart in Tables \ref{tab:res_rcv}
and \ref{tab:res_jrc}). A possible reason for this might be the fact
that the amount of training documents available in our experimental
setting is insufficient for learning a meaningful composition
function. A further possible reason might be the fact that, in
classification by topic, the mere presence or absence of certain
predictive words captures most of the information useful for
determining the correct class labels, while the information
conveyed 
by word order is less useful, or too difficult to capture. In future
work it might thus be interesting to test the Recurrent VGF on tasks
other than classification by topic.

\blue{Another aspect that jumps to the eye is that the relative
improvements brought about by the addition of WCEs tend to be larger
in JRC-Acquis than in RCV1/RCV2 (in which the presence of WCEs is
sometimes detrimental). This is likely due to the fact that JRC-Acquis
has more classes, something that ends up enriching the representations
of WCEs. Somehow surprisingly, though, the best configuration is one
not equipped with WCEs (and this happens also for JRC-Acquis). This
might be due to a redundancy of the information captured by WCEs with
respect to the information already captured in the other views. In the
future, it might be interesting to devise ways for distilling the
novel information that a VGF could contribute to the already existing
views, and discarding the rest during the aggregation phase.}

\begin{table}[t]
  \centering
  \begin{tabular}{lcccc}
    \toprule
    Method & $F_{1}^{M}$ & $F_{1}^{\mu}$ & $K^{M}$ & $K^{\mu}$\\
    \midrule
    \gfun-\vggM & .439 $\pm$ .072\psdag & .717 $\pm$ .067\psdag & .450 $\pm$ .091\psdag & .692 $\pm$ .071\psdag \\
    \gfun-\vggMW & .431 $\pm$ .086\psdag & .731 $\pm$ .064\psdag & .411 $\pm$ .102\psdag & .665 $\pm$ .081\psdag \\
    \blue{\gfun-\vgbgM} & .566 $\pm$ .065\psdag & .810 $\pm$ .040\psdag & .559 $\pm$ .083\psdag & .774 $\pm$ .050\psdag \\
    \gfun-\vgbgMW 	& .581 $\pm$ .064\psdag & .813 $\pm$ .039\psdag & .582 $\pm$ .080\sdpsdag & .794 $\pm$ .049\psdag \\
    \gfun-\vgxgMW 	 	& .527 $\pm$ .060\psdag & .788 $\pm$ .042\psdag & .531 $\pm$ .073\psdag & .777 $\pm$ .049\psdag \\
    \gfun-\vgxbgM & \textbf{.603} $\pm$ .066\psdag & \textbf{.826} $\pm$ .038\psdag & \textbf{.601} $\pm$ .077\psdag & \textbf{.811} $\pm$ .046\psdag \\
    \gfun-\vgxbgMW 	& .581 $\pm$ .059\psdag & .815 $\pm$ .037\psdag & .583 $\pm$ .074\sdpsdag & .799 $\pm$ .047\psdag \\
    \bottomrule
  \end{tabular}
  \caption{Cross-lingual text classification results on RCV1/RCV2
  dataset. Tests of statistical significance are performed against the
  best results found in Table \ref{tab:res_rcv}.}
  \label{tab:res_rcv:rec}
\end{table}

\begin{table}[t]
  \centering
  \begin{tabular}{lcccc}
    \toprule
    Method & $F_{1}^{M}$ & $F_{1}^{\mu}$ & $K^{M}$ & $K^{\mu}$\\
    \midrule
    \gfun-\vggM & .225 $\pm$ .074\psdag & .379 $\pm$ .096\psdag & .234 $\pm$ .076\psdag & .354 $\pm$ .096\psdag \\
    \gfun-\vggMW & .314 $\pm$ .019\psdag & .488 $\pm$ .022\psdag & .281 $\pm$ .020\psdag & .393 $\pm$ .024\psdag \\
    \blue{\gfun-\vgbgM} & .390 $\pm$ .027\psdag & .561 $\pm$ .021\psdag & .358 $\pm$ .027\psdag & .466 $\pm$ .021\psdag \\
    \gfun-\vgbgMW 	& .470 $\pm$ .017\psdag & .598 $\pm$ .013\psdag & .472 $\pm$ .020\psdag & .564 $\pm$ .018\psdag \\
    \gfun-\vgxgMW 	& .418 $\pm$ .011\psdag & .569 $\pm$ .008\psdag & .423 $\pm$ .012\psdag & .528 $\pm$ .010\psdag \\
    \gfun-\vgxbgM & \textbf{.501} $\pm$ .016\psdag & \textbf{.634} $\pm$ .011\psdag & \textbf{.501} $\pm$ .020\psdag & \textbf{.595} $\pm$ .016\psdag \\
    \gfun-\vgxbgMW & .483 $\pm$ .011\psdag & .615 $\pm$ .008\psdag & .482 $\pm$ .014\psdag & .577 $\pm$ .011\psdag \\

    \bottomrule
  \end{tabular}
  \caption{As Table~\ref{tab:res_rcv:rec}, but using JRC-Acquis
  instead of RCV1/RCV2.}
  \label{tab:res_jrc:rec}
\end{table}



\section{Related work}\label{sec:relwork}

\noindent The first published paper on CLTC is~\cite{Bel:2003dk}; in
this work, as well as in~\cite{Adeva:2005zx}, the task is tackled by
means of a bag-of-words representation approach, whereby
the texts are represented as standard vectors
of length $|\mathcal{V}|$, with $\mathcal{V}$ being the union of the
vocabularies of the different languages. Transfer is thus achieved
only thanks to features shared across languages, such as proper names.

Years later, the field started to focus on methods originating from
\textit{distributional semantic models} (DSMs)~\cite{Schutze:1993jr,
Sahlgren:2006yq, Landauer:1997if}. These models are based on the
so-called ``distributional hypothesis'', which states that similarity
in meaning results in similarity of linguistic distribution
\cite{Harris:1954fe}. Originally, these models \cite{Dumais:1997si,
Mimno:2009hp} made use of \textit{latent semantic analysis}
(LSA)~\cite{Deerwester:1990qc},
which factors a term co-occurrence matrix 
by means of low-rank approximation techniques such as SVD, resulting
in a matrix of principal components, where each dimension is linearly
independent of the others.
The first examples of cross-lingual representations
were proposed during the '90s. Many of these early works relied on
abstract linguistic labels, such as those from \textit{discourse
representation theory} (DRT) \citep{Kamp:1988pj}, instead of on purely
lexical features \citep{Aone:1993sy, Schultz:2001xr}. Early approaches
were based on the construction of high-dimensional context-counting
vectors where each dimension represented the degree of co-occurrence
of the word with a specific word in one of the languages of
interest. However, these original implementations of DSMs required to
explicitly compute the term co-occurrence matrix, making these
approaches unfeasible for large amounts of data.

A seminal work is that of~\citet{Mikolov:2013no}, who first noticed
that continuous word embedding spaces exhibit similar topologies
across different languages, and proposed to exploit this similarity by
learning a linear mapping from a source to a target embedding space,
exploiting a parallel vocabulary for providing anchor points for
learning the mapping. This has spawned several studies on
cross-lingual word embeddings~\cite{Faruqui:2014fk, Artetxe:2016ck,
Xing:2015fm}; however, all these methods relied on external manually
generated resources (e.g., multilingual seed dictionaries, parallel
corpora, etc.). This is a severe limitation, since the quality of the
resulting word embeddings (and the very possibility to generate them)
relies on the
availability, and the quality, of these external
resources~\citep{Levy:2015lo}.
%
%

Machine Translation (MT) represents a natural direct solution to CLTC
tasks. Unfortunately, when it comes to low-resource languages, MT
systems may be either not available or not sufficiently
effective. Nevertheless, the MT-based approach will presumably become
more and more viable as the field of MT progresses:
recently,~\citet{Isbister:2021rl} have shown evidence that relying on
MT in order to translate documents from low-resource languages to
higher-resource languages (e.g., English) for which state-of-the-art
models are available, is indeed preferable to multilingual solutions.
Pre-trained word-embeddings~\citep{Bengio:2003sf, Mikolov:2013,
Pennington:2014eu} have been a major breakthrough for NLP and have
become a key component of most natural language understanding
architectures. As of today, many methods developed for CLTC rely on
pre-trained \textit{cross-lingual word embeddings}
\cite{Mikolov:2013no, Artetxe:2017xa, Smith:2017qj, Conneau:2018bv}
(for a more in-depth review on the subject
see~\cite{Ruder:2019ud}). These embeddings strive to map
representations from one language to the other via different
techniques (e.g., Procrustes alignment), thus representing different
languages in different, but aligned, vector spaces. For
example,~\cite{Chen:2019jv, Zhang:2020hh} exploit aligned
word embeddings in order to successfully transfer knowledge from one
language to another. The approach proposed in~\cite{Chen:2019jv} is a
hybrid parameter-based / feature-based method to CLTC, in which a set
of convolutional neural networks is trained on both source and target
texts, encoded via aligned word representations (namely,
MUSEs~\cite{Conneau:2018bv}) while sharing kernel parameters to better
identify the common features across different languages. Furthermore,
the authors insert in the loss function a regularisation term based on
maximum mean discrepancy~\citep{Gretton:2012rr} in order to encourage
representations that are domain-invariant.

Standard word embeddings have recently been called \textit{static} (or
\textit{global}) representations. This is because they do not take
into account the context of usage of a word, thus allowing only a
single context-independent representation for each word; in other
words, the different meanings of polysemous words are collapsed into a
single representation. By contrast, \textit{contextual} word
embeddings~\citep{Melamud:2016qv, McCann:2017bv, Peters:2018zt,
Devlin:2019xx} associate each word occurrence with a representation
that is a function of the entire sequence in which the word
appears. Before processing each word with the ``contextualising''
function, tokens are mapped to a primary static word representation by
means of
a \textit{language model}, typically implemented by a transformer
architecture previously trained on large quantities of textual data.
This has yielded a shift in the way we operate with embedded
representations, from a setting in which pre-trained embeddings were
used to initialize the embedding layer of a deep architecture that is
later fully trained, to another in which the representation of words,
phrases, and documents, is carried out by the transformer; what is
left for training entails nothing more than learning a prediction
layer, and possibly fine-tuning the transformer for the task at hand.

Such a paradigm shift has fuelled the appearance of models developed
(or adapted) to deal with multilingual scenarios. Current multilingual
models are large architectures directly trained on several languages
at once, i.e., are models in which multilingualism is imposed by
constraining all languages to share the same model
parameters~\cite{Lample:2019kt, Eisenschlos:2019zl, Devlin:2019xx}.
Given their extensive multilingual pre-training, such models are
almost ubiquitous components of CLTC solutions.

\blue{For example, \citet{Zhang:2020hh} rely on pre-trained
multilingual \label{line:margin-aware} BERT in order to extract
word representations aligned between the source and the target
language. In a multitask-learning fashion, two identical-output
(linear) classifier sare set up: the first is optimized on the source
language via cross-entropy loss, while the second (i.e., the auxiliary
classifier) is instead set to maximize the \textit{margin disparity
discrepancy}~\citep{Zhang:2019as}. This is achieved by driving the
auxiliary classifier to maximally differ (in terms of predictions)
from the main classifier when applied to the target language, while
returning similar predictions on the source language.}


\citet{Guo:2020ng} tackle mono-lingual TC by exploiting multilingual
data. They do so by using a contrastive learning loss as applied to
Chinese BERT, a pre-trained (monolingual) language model. Then a
unified model, which is composed of two trainable pooling layers and
two auto-encoders, is trained on the union of the training data coming
from both the source and the target languages. It is important to note
that such a parameter-based approach requires parallel training data
in order to successfully train the auto-encoders (i.e., so that they
are able to create representations shared between the source and the
target languages).

\citet{Karamanolakis:2020nz} propose a parameter-based approach. They
first train a classifier on the source language, and then leverage the
learned parameters of a set of $b$ ``seed'' words to initialize the
target language model
(where $b$ refers to the number of words that can be translated to the
target language by a translation oracle). Subsequently, this model is
used as a \textit{teacher}, in knowledge-distillation fashion, to
train a \textit{student} classifier which is able to generalize beyond
the $b$ words transferred from the source classifier to the target
classifier.

\blue{\citet{wang_cross-lingual_2021} leverage graph convolutional
\label{line:gnn} networks (GCNs) to integrate heterogeneous
information within the task. They create a graph with the help of
external resources such as a machine translation oracle and a
POS-tagger. In the constructed graph, documents and words are treated
as nodes, and edges are defined according to different relations, such
as part-of-speech roles, semantic similarity, and document
translations. Documents and words are connected by their
co-occurrences, and the edges are labelled with their respective
POSs. Document-document edges are also defined according to
document-document similarity, as well as between translation
equivalents. Once the heterogeneous cross-lingual graph is
constructed, GCNs are applied in order to calculate higher-order
representations of nodes with aggregated information. Finally, a
linear transformation is applied to the document components in order
to compute the prediction scores.}

\blue{\citet{van_der_heijden_multilingual_2021} demonstrates the
effectiveness \label{line:meta-learning} of meta-learning
approaches to cross-lingual text classification. Their goal is to
create models that can adapt to new domains rapidly from few training
examples. They propose a modification to MAML (Model-Agnostic
Meta-Learning) called ProtoMAMLn. MAML is a meta-learning approach
that optimises the base learner on the so-called ``query set'' (i.e.,
in-domain samples) after it has been updated on the so-called
``support set'' (that is, out-of-domain samples). ProtoMAMLn is an
adaptation of ProtoMAML, where prototypes (computed by ``Prototypical
Network'' \citep{Snell:2017dz}) are also L2-normalized.}

Unlike our system, all the previously discussed approaches are
designed to deal with a single source language only. Nevertheless, as
we have already specified in Section \ref{sec:introduction}, a
solution designed to natively deal with multiple sources would be
helpful. A similar idea is presented in \cite{Chen:2019kn},
where the authors propose a method that relies on an initial
multilingual representation of the document constituents. The model
focuses on learning, on the one hand, a private (invariant)
representation via an adversarial network, and on the other one, a
common (language-specific) representation via a mixture-of-experts
model. We do not include the system of \cite{Chen:2019kn} as a
baseline in our experiments since it was designed to dealing with
single-label problems.


\section{Conclusions}
\label{sec:conclusions}

\noindent We have presented Generalized Funnelling (\gfun), a novel
hierarchical learning ensemble method for heterogeneous transfer
learning, and we have applied it to the task of cross-lingual text
classification. \gfun\ is an extension of Funnelling (\fun), an
ensemble method where 1st-tier classifiers, each working on a
different and language-dependent feature space, return a vector of
calibrated posterior probabilities (with one dimension for each class)
for each document, and where the final classification decision is
taken by a meta-classifier that uses this vector as its input, and
that can thus exploit class-class correlations. \gfun\ extends \fun\
by allowing 1st-tier components to be arbitrary view-generating
functions, i.e., language-dependent functions that each produce a
language-agnostic representation (``view'') of the document. In the
instance of \gfun\ that we have described here,
\blue{\label{line:summary}for each document} the meta-classifier
receives as input a vector of calibrated posterior probabilities (as
in \fun) aggregated to other embedded representations
\blue{\label{line:summary2}of the document} that embody other types of
correlations, such as word-class correlations (as encoded by
``word-class embeddings''), word-word correlations (as encoded by
``multilingual unsupervised or supervised embeddings''), and
correlations between contextualized words (as encoded by multilingual
BERT). In experiments carried out on two large, standard datasets for
multilingual multilabel text classification, we have shown that this
instance of \textsc{gFun} substantially improves over \fun, and over
other strong baselines such as multilingual BERT itself. An additional
advantage of \gfun\ is that it is much better suited to zero-shot
classification than \fun, since in the absence of training examples
for a given language, views of the test document different from the
one generated by a trained classifier can be brought to bear.
 
Aside from its very good classification performance, \gfun\ has the
advantage
of having a ``plug-and-play'' character, since it allows arbitrary
types of view-generating functions to be plugged into the
architecture. A common characteristic in recent CLTC solutions is to
leverage some kind of available, pre-trained cross- or multilingual
resource; nevertheless, to the best of our knowledge, a solution
trying to capitalise on multiple different (i.e., heterogeneous)
resources has not yet been proposed. Furthermore, most approaches aim
at improving the performance on the target language by exploiting a
single source language (i.e., they are single-source approaches).
In this, \gfun\ differs from the discussed solutions since (i) it
fully capitalises on multiple, heterogeneous available resources, (ii)
\blue{\label{line:capable}while capable in principle to deal with
single-source settings, it is especially designed to be deployed in
multi-source settings} and (iii) it is an
``everybody-helps-everybody'' solution, meaning that each
language-specific training set contributes to the classification of
all the documents, irrespectively of their language, and that all the
languages benefit from the inclusion of other languages in the
training phase (in other words, all the languages play both the role
of source and target at the same time).


Finally, we note that \gfun\ is a completely general-purpose
heterogeneous transfer learning architecture, and its application
(once appropriate VGFs are deployed) is not restricted to
cross-lingual settings, or even to scenarios where text is
involved. Indeed, in our future work we plan to test its adequacy to
cross-media applications, i.e., situations in which the domains across
which knowledge is transferred are represented by different media
(say, text and images).





\section*{Acknowledgements}\label{sec:acknowledgements}

\noindent The present work has been supported by the
\textsf{ARIADNEplus} project, funded by the European Commission (Grant
823914) under the H2020 Programme INFRAIA-2018-1, by the
\textsf{SoBigdata++} project, funded by the European Commission (Grant
871042) under the H2020 Programme INFRAIA-2019-1, and by the
\textsf{AI4Media} project, funded by the European Commission (Grant
951911) under the H2020 Programme ICT-48-2020. The authors' opinions
do not necessarily reflect those of the European Commission.





\end{document}